\documentclass[11pt,a4paper]{article}
\usepackage{times,latexsym}
\usepackage{url}
\usepackage[T1]{fontenc}
\usepackage{subcaption}
\usepackage{multirow}
\usepackage{adjustbox}
\usepackage{array}
\usepackage{booktabs}
\usepackage{graphicx}
\usepackage{cancel}
\usepackage{mathptmx}
\usepackage{amssymb}
\usepackage{xfrac}
\usepackage{amsmath}
\usepackage{amsfonts}
\usepackage{amsthm}
\usepackage{makecell}

\usepackage{fancyvrb}

\usepackage[most]{tcolorbox}
\newtcolorbox{exampleblock}[1]{colback=green!10!white,colframe=green!60!black,title=#1}

\newcolumntype{L}[1]{>{\raggedright\arraybackslash}p{#1}}

\usepackage{pifont}

\usepackage{siunitx}
\usepackage[acceptedWithA]{tacl2021v1}
\usepackage{xspace,mfirstuc,tabulary}
\usepackage{cleveref}

\newif\iftaclinstructions
\taclinstructionsfalse %
\iftaclinstructions
\renewcommand{\confidential}{}
\renewcommand{\anonsubtext}{(No author info supplied here, for consistency with
TACL-submission anonymization requirements)}
\newcommand{\instr}
\fi

\iftaclpubformat %

\else

\fi

\newcommand{\rparagraph}[1]{\vspace{1.2mm}\noindent\textbf{#1.}}

\newcommand{\fire}[0]{\includegraphics[width=.015\textwidth]{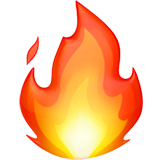}}
\newcommand{\snowflake}[0]{\includegraphics[width=.015\textwidth]{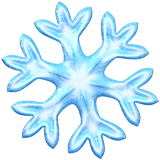}}
\newcommand{\one}[0]{\includegraphics[width=.02\textwidth]{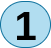}}
\newcommand{\two}[0]{\includegraphics[width=.02\textwidth]{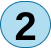}}

\title{Supervised In-Context Fine-Tuning for \\ Generative Sequence Labeling}

\author{David Dukić$^{1}$ \and Goran Glavaš$^{2}$  \and Jan Šnajder$^1$  \\ $^1$University of Zagreb Faculty of Electrical Engineering and Computing, TakeLab \\ $^2$CAIDAS, University of Würzburg, Germany \\ $^{1,2}$\texttt{name.surname@\{fer.hr, uni-wuerzburg.de\}}} 

\date{}

\begin{document}

\pagestyle{plain}
\pagenumbering{arabic}
\setcounter{page}{1}
\thispagestyle{plain}

\maketitle

\begin{abstract}
    Sequence labeling (SL) tasks, where labels are assigned to tokens, are abundant in NLP (e.g., named entity recognition and aspect-based sentiment analysis). Owing to the intuition that they require bidirectional context, SL tasks are commonly tackled with encoder-only models. Recent work also shows that removing the causal mask in fine-tuning enables decoder-based LLMs to become effective token classifiers. Less work, however, focused on (supervised) generative SL, a more natural setting for causal LLMs. Due to their rapid scaling, causal LLMs applied to SL are expected to outperform encoders, whose own development has stagnated. In this work, we propose supervised in-context fine-tuning (SIFT) for generative SL. SIFT casts SL tasks as constrained response generation, natural to LLMs, combining in-context learning (ICL) from demonstrations with supervised fine-tuning. SIFT considerably outperforms both ICL and decoder-as-encoder fine-tuning baselines on a range of standard SL tasks. We further find that although long context hinders the performance of generative SL in both ICL and SIFT, this deficiency can be mitigated by removing the instruction, as instructions are shown to be largely unnecessary for achieving strong SL performance with SIFT. Our findings highlight strengths and limitations of SL with LLMs, underscoring the importance of a response-based generative task formulation for effective SL performance.

\end{abstract} %
\section{Introduction}

\begin{figure}[!htb]
    \centering
    \includegraphics[width=1.0\linewidth]{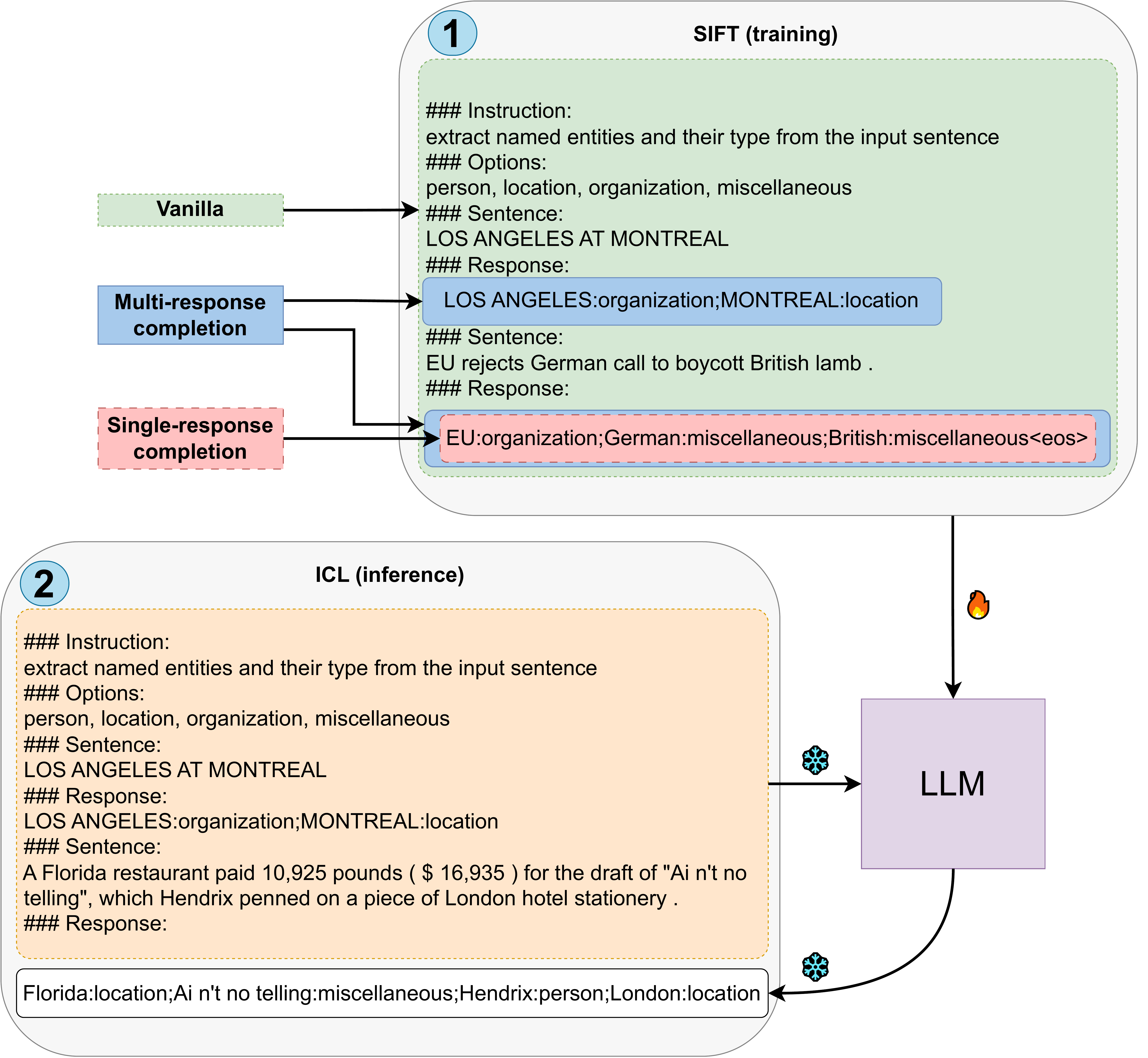}
    \caption{\one \hspace{0.015em} Supervised in-context fine-tuning (SIFT) for sequence labeling tasks with three different strategies for generative fine-tuning (\fire) with in-context demonstrations:     
    (a) \textit{vanilla}: causal language modeling (CLM) carried out on all prompt tokens; (b) single-response completion (\textit{SRC}): CLM on the response tokens of the last instance; and (c) multi-response completion (\textit{MRC}): CLM on response tokens of all demonstration instances and last, target instance. \two \hspace{0.015em} In-context learning (ICL) with constrained decoding at inference (\snowflake).  
    }
    \label{fig:clm_strategies}
\end{figure}

Sequence labeling (SL) falls under a class of natural language understanding (NLU) tasks that require deep linguistic comprehension of the underlying text to deliver accurate results. Unlike most NLU tasks, such as text classification and question answering, SL requires assigning a label to each token within a sequence, making it inherently more complex while simultaneously offering greater versatility and convenience for numerous downstream applications. Outputs of models trained for SL are used to populate knowledge bases \citep{mesquita-etal-2019-knowledgenet,radevski2023linking}, power news and social media monitoring systems \citep{osborne-etal-2014-real,dukic2024closed}, create strong virtual assistants \citep{razumovskaia-etal-2022-natural,razumovskaia-etal-2023-transfer}, and facilitate research in social sciences by automating large-scale text analysis \citep{pado-etal-2019-sides,klamm-etal-2023-kind,dukic2024takelab}. %

Previously, the models of choice for SL tasks were pre-trained encoder-only transformers such as BERT \citep{devlin-etal-2019-bert} and RoBERTa \citep{liu2019roberta}. However, with the rise in the number of parameters and pre-training data, decoder-only large language models (LLMs) have shifted the landscape in their favor. Not only can they be adapted with supervised fine-tuning (SFT) to solve NLU tasks in a generative manner, but they also showed the ability to solve novel tasks relying on in-context learning (ICL) \cite{brown2020language}. The rapid scaling of causal LLMs suggests that this approach to SL may offer a faster trajectory for performance gains compared to encoder-based models, whose scaling has effectively plateaued. Furthermore, decoder-only LLMs have proven effective when post-hoc repurposed as encoders for SL tasks, consistently outperforming instruction-tuned decoders and strong encoder-only baselines \cite{li2023label,dukic-snajder-2024-looking}. 
SFT and ICL possess distinct strengths and limitations. On the one hand, SFT  of LLMs generally yields strong task-specific performance, but requires a task-specific fine-tuning procedure that updates a subset or all of LLM's parameters. On the other hand, ICL does not require parameter updates, but provides labeled instances, commonly referred to as \textit{demonstrations}, to the LLM as part of the input context, i.e., in the prompt. ICL abilities, which emerge only in larger-scale LLM pre-training \citep{brown2020language,wies2023learnability}, allow the model to perform tasks unseen in pre-training and infer (``learn'') from the provided task description (i.e., instruction) and demonstrations \citep{min-etal-2022-rethinking}. While ICL is more flexible (i.e., no updates to the underlying LLM) and computationally cheaper than SFT, it typically cannot reach the performance of task-specific SFT:   
fine-tuning beats long-context ICL once presented with enough training examples \citep{bertsch-etal-2025-context}. These observations suggest that an effective SL strategy may integrate ICL and SFT by leveraging their complementary strengths.

In this work, we explore the potential of \textit{generative sequence labeling with LLMs} and introduce response adaptation strategies for supervised in-context fine-tuning (SIFT), a framework that unifies SFT and ICL for SL, as shown in 
\Cref{fig:clm_strategies}. The SIFT framework gives rise to three sensible strategies for generative fine-tuning with in-context demonstrations: (1) \textit{vanilla}, where standard causal language modeling (CLM) is applied to the entire prompt (i.e., the loss is computed on all tokens), (2) \textit{single-response completion (SRC)}, where the model predicts only the response tokens of the ``target'' instance (i.e., the last response, excluding demonstrations), and (3) \textit{multi-response completion (MRC)}, where the model predicts the response tokens for both the demonstrations and the final, ``target'' instance. At inference time, we carry out ICL with constrained token generation using the model adapted with SIFT.

Our experiments, covering four well-established SL tasks (named entity recognition, aspect-based argument mining, slot labeling in dialogs, and semantic role labeling) and five modern LLMs, demonstrate the effectiveness of our response adaptation strategies for SIFT: SIFT not only outperforms standard SFT and ICL, but also the competitive decoder-as-encoder fine-tuning \cite{dukic-snajder-2024-looking, behnamghaderllm2vec}.  
We obtain the best results with the multi-response completion (MRC) training strategy, demonstrating that generative fine-tuning benefits from jointly computing the loss over multiple in-context examples.  
We further find that our SIFT strategies, as well as plain ICL, struggle with long contexts, inherent to the generative formulation of SL tasks. 
Finally, we show that SIFT removes the LLMs' sensitivity to the instruction: we obtain similar performance with different instructions as well as without the instruction altogether. Our findings push the performance boundaries and lend themselves to reliable best-practice recommendations with respect to generative SFT for SL tasks. 
\begin{figure*}[t!]
    \centering
    \includegraphics[width=\linewidth]{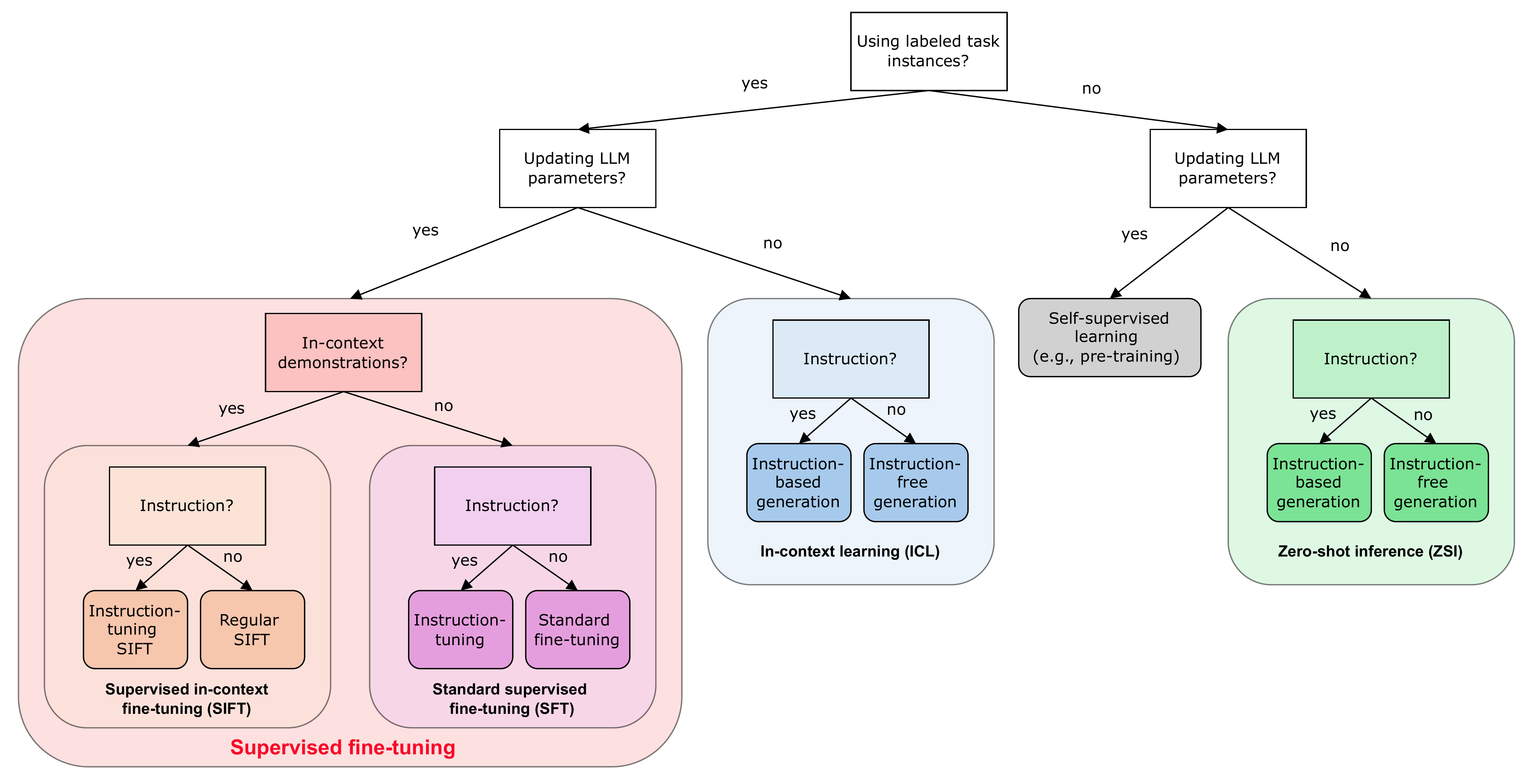}
    \caption{Supervised in-context fine-tuning (SIFT; in orange box) as a task-specific learning paradigm for LLMs, in relation to (standard) supervised fine-tuning (SFT; in purple box) and in-context learning (ICL; in blue box). For completion, zero-shot inference (no labeled instances) is shown in the green box.}
    \label{fig:taxonomy} %
\end{figure*}

\section{Supervised In-Context Fine-Tuning}

Before detailing the SIFT response adaptation strategies, we situate SIFT within the broader landscape of LLM learning approaches. The decision tree in \Cref{fig:taxonomy} positions SIFT in relation to SFT and ICL and demonstrates that SIFT effectively integrates the two.

Both SFT and ICL rely on labeled task-specific examples. ICL leverages labeled examples, called demonstrations, by incorporating them into the prompt without altering the model’s parameters in any way. SFT, in contrast, uses labeled examples as actual training instances based on which a loss function is computed and the model's parameters are updated. While SFT can leverage a virtually unlimited number of labeled instances, ICL is constrained by the LLM's context size, limiting the number of demonstrations. With earlier models, ICL was limited to ``few-shot ICL'' \cite{mosbach2023few}, as their smaller context could only accommodate a ``few'' demonstrations. More recent models, however, with context lengths in the tens or hundreds of thousands of tokens, now enable many-shot ICL \cite{agarwal2024many}.

In both ICL and SFT, one can start the context with an instruction, that is, a natural language description of the task. This, intuitively, makes more sense if the underlying model is an instruction-tuned LLM, that is, a model that has been fine-tuned on instruction-response pairs spanning many diverse tasks \cite{mishra-etal-2022-cross}, although instruction following may also emerge without instruction-tuning \cite{hewitt2024instruction}.

Supervised in-context fine-tuning (SIFT)---shown in the orange box in the decision tree in \Cref{fig:taxonomy}---can be seen as a hybrid between SFT and ICL: we update the model's parameters (as in SFT) based on the labeled examples in the context, but the context, besides the last, ``target'' instance, contains additionally in-context demonstrations (as in ICL). Like both standard SFT and ICL, SIFT may or may not include the task instruction at the beginning of the prompt. 

What remains to be defined for a concrete SIFT training run is the actual learning objective. Since we focus on decoder-based LLMs, the objective has to be generative, i.e., token prediction. However, the question of which tokens in the context to predict, including the (optional) instruction, demonstrations, and the last target instance, remains unanswered. In this work, we investigate three different SIFT training strategies for SL tasks, described in detail in the next section.           

\section{SIFT for Sequence Labeling}
\label{sec:sift}

We propose a SIFT framework for SL tasks, investigating different fine-tuning objectives and comparing SIFT empirically to standard SFT and ICL. %

A SIFT prompt consists of three parts: (1) the instruction, (2) the demonstrations, i.e., in-context examples with gold responses, and (3) the query: the final example. At training time, the query example is also coupled with the gold response, and as such it does not really differ from the in-context demonstrations (it is essentially the last demonstration); at inference time, in a standard ICL setup, we have the demonstrations and the query example and the model generates its response; in-context demonstrations still consist of the example and gold response. Note that removing in-context demonstrations (i.e., part (2)) reduces SIFT to standard SFT. We provide both training and inference templates for both SIFT (with in-context demonstrations) and standard SFT (no in-context demonstrations) in \Cref{tab:token_div}. Additionally, we also evaluate both SIFT and standard SFT without the instruction in the prompt. 

Since we are in the realm of generative decoder-only LLMs, our SIFT objectives are generative, i.e., based on token prediction: (1) \textit{vanilla} SIFT trains on all tokens, i.e., amounts to standard CLM on the entire prompt; (2) \textit{single-response completion} (SRC) predicts only the tokens of the response to the query (i.e., the last response)---we will denote this set of query response tokens with $\mathit{QR}$, and (3) \textit{multi-response completion} predicts all tokens in responses of all demonstrations as well as the query response---we will denote the set of all tokens from responses of all in-context demonstrations with $\mathit{DR}$.  
\Cref{fig:clm_strategies} illustrates all three SIFT strategies. 
SIFT models require in-context demonstrations for both training and inference. Unless specified otherwise, the number of demonstrations at inference matches the number used for training.

\rparagraph{Vanilla SIFT} 
This is the standard CLM objective, as used in LLM pre-training. 
Formally, we define the vanilla training loss over every token of the training input sequence $\mathbf{t}=(t_1, \dots, t_N)$:
\[
L_V(\mathbf{t};\mathbf{\theta}) = - \sum_{i=1}^{N}\log P(t_i | t_{<i};\mathbf{\theta})
\]
\noindent where $\mathbf{\theta}$ are trainable model parameters and $t_i$ is the $i$-th token in the training sequence out of $N$ tokens, conditioned on preceding context $t_{<i}$.

\rparagraph{Single-Response Completion}
This strategy adapts the decoder by masking out all tokens except those in $\mathit{QR}$ from the loss computation, steering the model to generate valid responses to an example query. %
Formally, for each training input sequence $\mathbf{t}$, we define the SRC loss over a subset of training tokens:
\[
L_{SRC}(\mathbf{t};\mathbf{\theta}) = - \sum_{i=1}^{N} \delta_i \log P(t_i | t_{<i};\mathbf{\theta}),
\]

\[
\delta_i =
\begin{cases} 
1 & \text{if } t_i \in \mathit{QR} \\
0 & \text{otherwise}
\end{cases}
\]

\noindent where $\mathbf{\theta}$ are trainable model parameters, $t_i$ is is the $i$-th token in the training sequence out of $N$ tokens, conditioned on preceding token context $t_{<i}$, and $\delta_i$ is the indicator function, which keeps the loss only for the $\mathit{QR}$ tokens.

\rparagraph{Multi-Response Completion}
Here we extend SRC to all responses in the context, i.e., we do not compute the CLM loss only for the query response $\mathit{QR}$ but also for the responses of all in-context demonstrations, i.e., all tokens in $\mathit{DR}$.    
This is meant to force the model to generate correct answers for multiple instances simultaneously, while at the same time using for each generated response all other instances as context. 
Formally, for each training input sequence $\mathbf{t}$, we define MRC loss as follows:
\[
L_{MRC}(\mathbf{t};\mathbf{\theta}) = - \sum_{i=1}^{N} \delta_i \log P(t_i | t_{<i};\mathbf{\theta}),
\]

\[
\delta_i =
\begin{cases} 
1 & \text{if } t_i \in \mathit{QR} \cup \mathit{DR} \\
0 & \text{otherwise}
\end{cases}
\]
\noindent where $\mathbf{\theta}$ are trainable model parameters, $t_i$ is is the $i$-th token in the training sequence out of $N$ tokens, conditioned on preceding token context $t_{<i}$, and $\delta_i$ is the indicator function, which holds true only for tokens in $\mathit{QR}$ and tokens in $\mathit{DR}$.
\section{Experimental Setup}

We compare the \textit{base} decoders with their instructed and dialogue-optimized variants (\textit{instruct}) as we increase the number of demonstrations during fine-tuning and inference. Further, we compare SIFT with standard SFT and ICL. We juxtapose these models with decoder-as-encoder baselines: unmasking from \citet{dukic-snajder-2024-looking} (causal mask (CM) is removed from all the layers) and the LLM2Vec approach from \citet{behnamghaderllm2vec}. Next, we compare the CLM strategies and underscore the performance differences. Finally, to observe how much decoders depend on the instruction, we experiment with instruction removal during fine-tuning and including variations of it during inference. We experiment on four SL tasks: named entity recognition (NER), aspect-based argument mining (ABAM), slot labeling in dialogs, and semantic role labeling (SRL). Training is done on the training portions, and evaluation on validation and test portions of the datasets (see \Cref{tab:dataset_stats}).

\subsection{Datasets}

\begin{table}
  \centering
  \small
  \setlength{\tabcolsep}{3.5pt}
  \begin{tabular}{@{}crrrrr@{}}
    \toprule
    \multicolumn{1}{c}{\multirow{1}{*}{\textbf{Dataset}}} & \multicolumn{1}{c}{\textbf{Train}} & \multicolumn{1}{r}{\textbf{Valid}} & \multicolumn{1}{r}{\textbf{Test}} & \multicolumn{1}{r}{\textbf{Total}} & \multicolumn{1}{r}{\textbf{\#Cl.}} \\
    \midrule
    OntoNotes v5.0 & \num{21244} & \num{5385} & \num{6000} & \num{32629} & 27 \\  
    CoNLL03 & \num{14041}  & \num{3250} & \num{3453} & \num{20744} & 4 \\
    NLU++ & \num{2152} & 309 & 619 & \num{3080} & 17 \\  
    AAC-MW & 670 & 95 & 193 & 958 & 12 \\  
    \bottomrule
  \end{tabular}
  \caption{Dataset statistics: the number of sentences per split, the total number of sentences, and the total number of classes. OntoNotes v5.0 was subsampled, while the remaining datasets were used with the original number of examples.}
  \label{tab:dataset_stats}
\end{table}

\rparagraph{NER} 
For NER, we choose CoNLL03 \citep{tjong-kim-sang-de-meulder-2003-introduction} dataset. We use the version available in \textit{Hugging Face Datasets} \citep{lhoest-etal-2021-datasets}, which follows the IOB2 sequence tagging scheme and provides predefined train, validation, and test splits. %

\rparagraph{ABAM}
This task was introduced by \citet{trautmann-2020-aspect} and boils down to the automatic detection and classification of argument aspects in a text. For this task, we leverage the Argument Aspect Corpus (AAC) with token-level annotations on four topics. We select the \textit{minimum wage (AAC-MW)} topic for experiments, which contains $12$ classes and an additional \textit{Other} class, which we exclude. Data splits were created using the code provided by the dataset authors.\footnote{\footnotesize{\url{https://github.com/Leibniz-HBI/argument-aspect-corpus-v1/blob/main/classification.py}}} The dataset was converted to IOB2 tags.

\rparagraph{Slot Labeling}
For slot labeling, we utilize the NLU++ dataset \citep{casanueva-etal-2022-nlu} and merge the \textit{banking} and \textit{hotels} domains, counting a total of $17$ classes. %
We create IOB2 tags for this dataset by mapping slot value offsets in the sequences with spaCy \citep{spacy} tokenization. %
The evaluation approach recommended by the authors was to use k-fold cross-validation. However, since this approach is resource-intensive in the context of LLM adaptation, we shuffle and randomly split the original folds into train/validation/test portions in a ratio of 70/10/20, to obtain fixed sets for measuring models' performance.

\rparagraph{SRL} 
We use the OntoNotes v5.0 dataset with the English v12 subset \citep{pradhan-etal-2013-towards}. Compared to the other datasets presented, this one stands out as the largest. We use predefined train, validation, and test portions. However, due to the large size of the original dataset, we downsampled both the training and validation sets to make experiments with multiple models computationally feasible. The test set was then adjusted to exclude examples with labels that were filtered from the training and validation sets, and then also downsampled. See \Cref{subsec:srl_processing} for details.

\subsection{Models}

We experiment with five open-weight LLMs in our experiments: Gemma-7B \citep{team2024gemma}, 
Llama2-7B \citep{touvron2023llama}, Llama3-8B, Llama3.1-8B \citep{dubey2024llama}, and Mistral-7B \citep{jiang2023mistral}. Models were selected based on the popularity on the \textit{Hugging Face Hub} (see \Cref{tab:sift_models} in \Cref{subsec:implementation_details}). By default, we use these decoder transformers with their pre-trained (C)LM-ing head.
For baseline experiments involving CM removal (and we remove future token masking from all decoder layers), we place a token classification head on top of the model's transformer body (see \Cref{subsec:cm_removal} for details). For LLM2Vec, we use pre-trained adapters released by the authors and follow their recommended procedure of fine-tuning only the classification head for sequence labeling tasks \citep{behnamghaderllm2vec}. See \Cref{subsec:llm2vec} for details.

\subsection{Optimization}

We apply QLoRA \cite{dettmers2023qlora} to query and value attention matrices in all decoder layers in all our fine-tuning experiments. We use the following hyperparameters: fixed rank $r=16$, scaling parameter $\alpha=16$, dropout probability of $p=0.1$, learning rate $2\mathrm{e}{-4}$, gradient accumulation, and a batch size of eight examples. We train the models with a paged 8-bit AdamW \citep{loshchilov2017decoupled} optimizer to handle the memory spikes. The parameters of AdamW are fixed to $\beta_1=0.9, \beta_2=0.95, \epsilon=1\mathrm{e}{-5}, \lambda=0.1$. We adopt the cosine annealing scheduling of the learning rate \citep{loshchilov2016sgdr}, set gradient clipping with a maximum value of $1.0$, and utilize gradient checkpointing. We carry out fixed-duration fine-tuning for five epochs over the training portions of the task datasets. 
For each fine-tuning experiment, we report the average performance from four different runs (i.e., with different random seeds).  
We refer the reader to \Cref{subsec:implementation_details} for optimization details and \Cref{subsec:special_tokens} for details on handling special tokens.

\subsection{Training and Evaluation} 

\rparagraph{Training} 
The training examples are constructed to adhere to the specified format with instruction, demonstrations, and query parts. We conduct training with $0$ (standard SFT), $1$, $5$, and $10$ demonstrations ($n$ shots) in the context part (see \Cref{tab:icl_sift_examples} in \Cref{sec:repl_details}). When the context includes one or more examples, we employ the SIFT setup (see \Cref{subsec:forming_training_examples} for details). The expected responses are formatted to adhere to a regular expression. This way, we train the models to learn to generate the spans and the classes simultaneously, saving on the total number of tokens required for the complete answer. We describe the generation scheme for the dataset that has four classes using the following regular expression:
\begin{center}
\scriptsize
\begin{BVerbatim}
NA|([^:;]+:(class1|class2|class3|class4);)*
[^:;]+:(class1|class2|class3|class4)
\end{BVerbatim}    
\end{center}
\noindent where we divide spans and their classes with colons, multiple extractions with semicolons, and direct the model to generate \texttt{NA} when no spans are to be extracted from the target instance.

\rparagraph{Evaluation} 
We evaluate on IOB2 tags with strict matching using the micro F1 score on all models. To ensure a fair evaluation consistent with models fine-tuned directly for SL, we heuristically map response spans of decoders trained with SFT and SIFT to IOB2 tags. We employ greedy span-based matching of predicted spans and their classes with input tokens, similar to \citet{wang2022instructionner} (see \Cref{subsec:evaluation_details} for details). We match the number of demonstrations in the context for training and evaluation, as this approach has proven to be the best on the validation set. The context for the higher number of shots was kept fixed for a fixed seed. We leverage the \textit{outlines} library \citep{willard2023efficient} for constrained generation to ensure the models follow the generation scheme specified with a regular expression. We utilize the \textit{vLLM} library \citep{kwon2023efficient} to accelerate generation. See \Cref{subsec:forming_evaluation_examples} for more details. 

\subsection{Setup for the Instruction Ablation}
To measure the effect of instruction on the performance of standard SFT and SIFT models, we first train the model for the task without the instruction and then evaluate using ICL with variations of the task instruction, matching the number of demonstrations used for training. The proposed variations are: (1) \textit{Vanilla}, where we evaluate with the usual task-specific instruction, (2) \textit{Permuted}, where we randomly permute the order of the tokens of the \textit{vanilla} instruction (with a fixed permutation seed) to observe if the model relies on the lexical presence of key words rather than the compositional structure of the instruction; and (3) \textit{Nonsense}, where we include a text snippet entirely unrelated to the task we trained the model for (see \Cref{subsec:instruction_variations}) to test whether the model depends on the instruction’s meaning to complete the task. %
\section{Results}

\subsection{Main Validation Set Results}

\begin{figure*}
\begin{center}
\includegraphics[width=\textwidth]{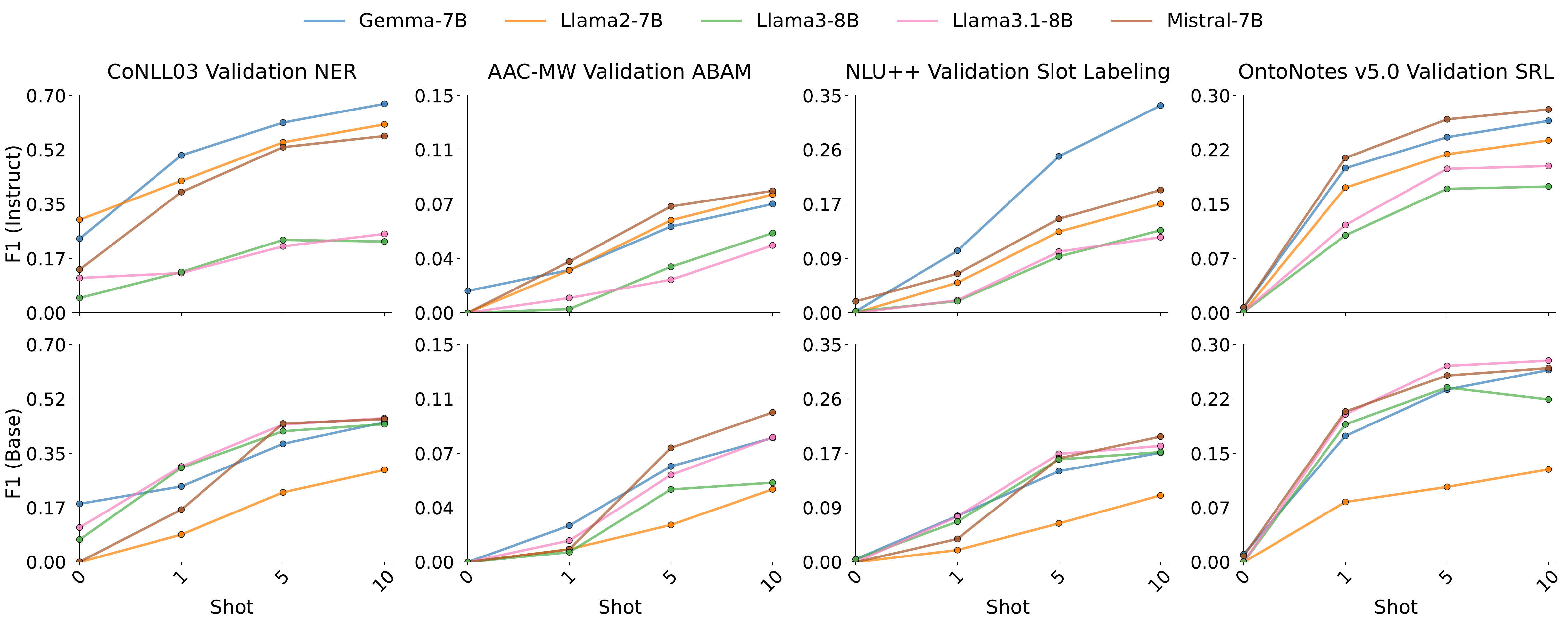}
\caption{Micro F1 scores using five \textit{base} and \textit{instruct} variants of decoders on ICL and for a varying number of shots. The x-axis shows the number of shots on an ordinal scale. The results are given for the validation set on four tasks (left to right), with top row plots corresponding to \textit{instruct} variants and bottom row plots corresponding to \textit{base} variants. All results are averages of four runs.}
\label{fig:icl_sl}
\end{center}
\end{figure*}

\rparagraph{ICL}
\Cref{fig:icl_sl} shows the validation set performance for ICL. We note an increasing F1 trend as we raise the number of shots. This trend is more pronounced for \textit{instruct} than \textit{base} variants, as \textit{instruct} models reach higher ICL performance per task. Gemma-7B-Instruct dominates the performance on the NER and slot labeling, while Mistral-7B reaches the best results on ABAM, and Mistral-7B-Instruct reaches the best results on the SRL. The most challenging task was ABAM. Generally, we observe fairly low performance for all tasks except NER, demonstrating that, in standard ICL, LLMs struggle with SL task completion across many classes. This is presumably because the model rarely observes instances from all classes within the demonstrations (max.~$10$ shots). Also, increasing the number of shots increases the context length, which is known to be detrimental for ICL \citep{10.1162/tacl_a_00638}. 

\begin{figure*}
\begin{center}
\includegraphics[width=\textwidth]{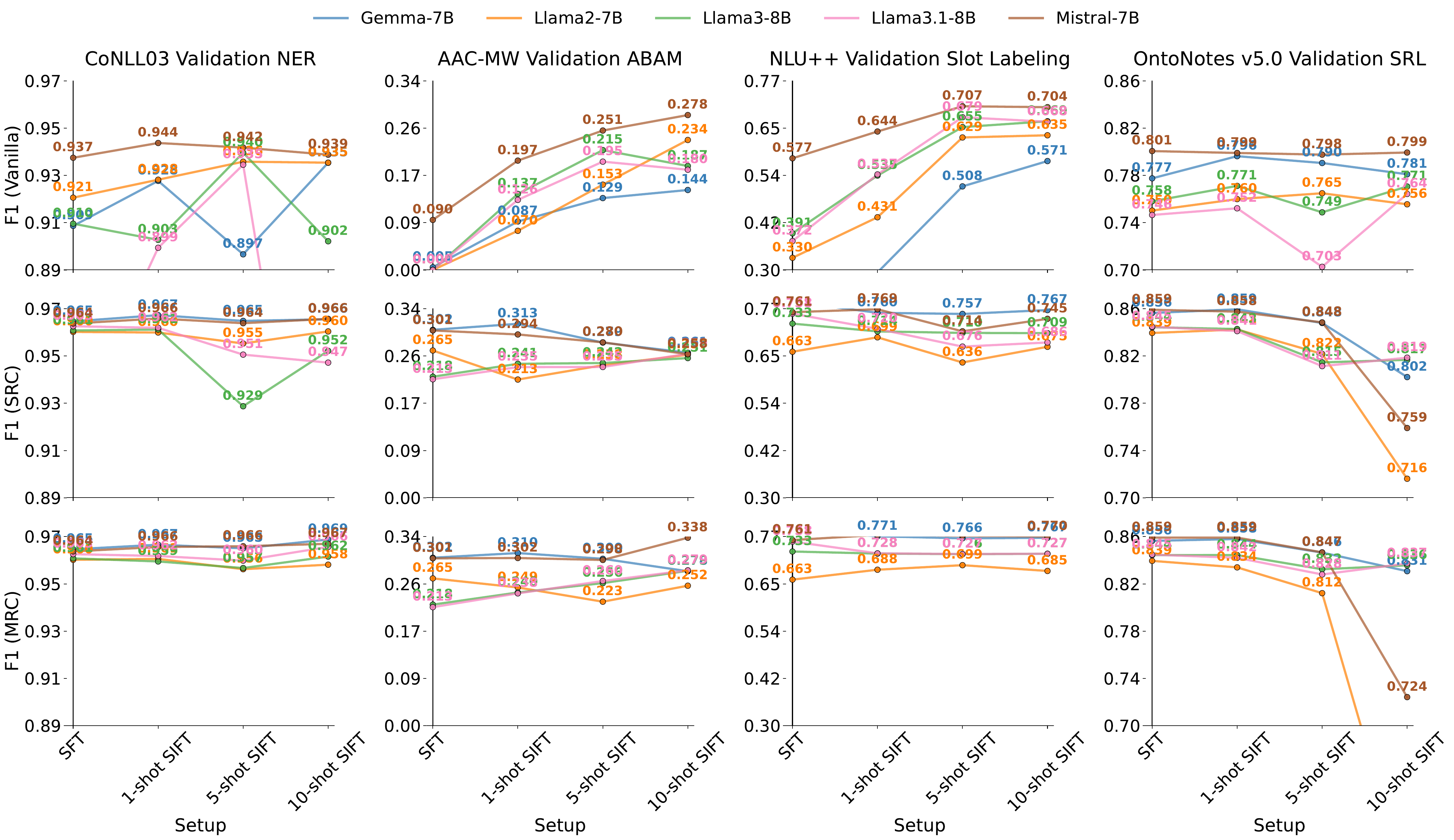}
\caption{Micro F1 scores for five \textit{base} variants of decoders on standard SFT and SIFT for a varying number of shots. The models are evaluated with the same number of shots in the context that they used for fine-tuning. The results are given for the validation set on four tasks (left to right) and for three CLM strategies (top to bottom). All results are averages of four runs. See \Cref{sec:complementary_res} for \textit{instruct} variants.}
\label{fig:it_sift_validation_sl}
\end{center}
\vspace{-1em}
\end{figure*}

\rparagraph{SIFT and Standard SFT}
We report the validation set results on four SL tasks in \Cref{fig:it_sift_validation_sl}, including the results for the three CLM strategies and \textit{base} decoders. \textit{Instruct} models performed worse (cf.~\Cref{fig:it_sift_validation_sl_instruct} in \Cref{sec:complementary_res}). Results reveal that vanilla CLM performs worse than SRC and MRC consistently across all datasets. However, increasing the number of demonstrations for vanilla CLM brings significant gains. On average, adding more demonstrations at least does not degrade performance. These gains are most apparent for the AAC-MW and NLU++ datasets. Consistent with vanilla CLM, we find that SRC and MRC also benefit from an increased number of demonstrations during fine-tuning and inference, except for the SRL task. Providing the model with more than one demonstration for the SRL task typically results in a decline in performance. All other tasks measure at least some benefits from more demonstrations in the context. NER shows fewer gains than ABAM and slot labeling. ABAM appears to be the most challenging task for the models to learn, presumably due to the small overall number of training examples. On average, the MRC strategy shows improvements more often than SRC with the increase in the number of shots---
aggregating the loss over multiple responses thus generally seems beneficial. MRC stabilizes the training and helps the model learn to leverage multiple responses in the context better than SRC. For both strategies, \textit{base} model variants score higher (winning in $96$ out of $160$ experiments) than the respective \textit{instruct} variants. Interestingly, Llama models perform worse than Gemma and Mistral in all experiments. Finally, when comparing standard SFT with SIFT, we observe that combinations of either (1) SRC + standard SFT or (2) MRC + SIFT yield the highest overall F1 score per model.

\rparagraph{ICL vs.~SIFT}
The performance gap between ICL and SIFT is substantial, ranging from around $0.1$ (ABAM) to more than $0.6$ (NER, slot labeling, and SRL) F1 points in favor of fine-tuning. \textit{Instruct} variants dominate ICL, while \textit{base} variants perform best in SIFT. Expectedly, ICL benefits more from a larger number of demonstrations than SIFT. With a large enough size of the fine-tuning dataset, the models saturate and do not benefit further from in-context demonstrations: This is consistent with prior work \citep{bertsch-etal-2025-context}. While more shots improve SIFT results, the benefits diminish with larger contexts (in the number of tokens), suggesting limits in the model's ability to effectively process extensive inputs \citep{10.1162/tacl_a_00638}.

\begin{table*}[!htb]
\centering
\small{
\setlength{\tabcolsep}{2pt}
\renewcommand{\arraystretch}{1.125} 
\begin{tabular}{l|lcccccccccccc}
\toprule
\multicolumn{2}{l}{\multirow{2}{*}{}} & \multicolumn{3}{c}{\textbf{CoNLL03}} & \multicolumn{3}{c}{\textbf{AAC-MW}} & \multicolumn{3}{c}{\textbf{NLU++}} & \multicolumn{3}{c}{\textbf{OntoNotes v5.0}} \\
\cmidrule(lr){3-5} \cmidrule(lr){6-8} \cmidrule(lr){9-11} \cmidrule(lr){12-14} 
\multicolumn{1}{l}{LM} & \multicolumn{1}{l}{Setup} & \textbf{ICL} & \textbf{SRC} & \textbf{MRC} & \textbf{ICL} & \textbf{SRC} & \textbf{MRC} & \textbf{ICL} & \textbf{SRC} & \textbf{MRC} & \textbf{ICL} & \textbf{SRC} & \textbf{MRC} \\
\midrule

\multirow{4}{*}{\rotatebox[origin=c]{90}{\shortstack{Gemma}}} &  0-shot/SFT 

& $24.5_{0.4}$  & $93.2_{0.3}$ & -- 
& $0.0_{0.0}$   & $35.9_{1.4}$ & -- 
& $0.0_{0.0}$   & $\underline{\mathbf{76.6}}_{0.7}$ & -- 
& $0.7_{0.1}$   &  $\mathbf{84.0}_{0.1}$ & -- \\

{} 

& 1-shot/SIFT 

& $48.4_{0.2}$ & $93.4_{0.3}$ & $93.6_{0.2}$ 
& $3.7_{1.3}$  & $\underline{\mathbf{37.2}}_{1.7}$ & $\mathbf{36.8}_{1.4}$ 
& $6.7_{3.9}$  & $76.0_{1.0}$ & $75.0_{0.6}$ 
& $17.4_{0.1}$ & $\underline{\mathbf{84.4}}_{0.1}$ & $84.0_{0.2}$ \\

{} 

& 5-shot/SIFT 

& $60.5_{0.4}$ & $93.4_{0.2}$ & $93.4_{0.1}$ 
& $6.5_{0.8}$  & $35.5_{0.5}$ & $35.7_{1.6}$ 
& $23.3_{3.7}$ & $75.2_{0.8}$ & $75.1_{0.7}$ 
& $21.3_{0.4}$ & $83.2_{0.2}$ & $82.7_{0.3}$ \\

{} 

& 10-shot/SIFT 

& $67.7_{0.5}$ & $\mathbf{93.4}_{0.2}$ & $\underline{\mathbf{93.8}}_{0.1}$ 
& $10.2_{1.2}$ & $36.6_{1.7}$ & $33.3_{1.0}$ 
& $30.8_{2.6}$ & $74.7_{0.7}$ & $\mathbf{75.6}_{1.7}$ 
& $23.7_{0.1}$ & $78.2_{9.0}$ & $81.3_{0.7}$ \\
\midrule

\multirow{4}{*}{\rotatebox[origin=c]{90}{\shortstack{Llama2}}}   &   0-shot/SFT 

& $28.2_{0.5}$ & $93.0_{0.1}$ & -- 
& $0.0_{0.0}$  & $\mathbf{29.9}_{1.2}$ & -- 
& $0.0_{0.0}$  & $66.9_{3.0}$ & -- 
& $0.1_{0.0}$  & $81.7_{0.3}$ & -- \\

{} & 1-shot/SIFT 

& $39.6_{0.3}$ & $92.9_{0.2}$ & $93.0_{0.2}$ 
& $3.8_{1.5}$  & $26.2_{1.3}$ & $\underline{\mathbf{31.2}}_{3.2}$ 
& $3.5_{2.1}$  & $\mathbf{68.6}_{2.1}$ & $67.8_{1.3}$ 
& $15.8_{0.1}$ & $\underline{\mathbf{82.2}}_{0.3}$ & $\mathbf{81.2}_{0.7}$ \\

{} & 5-shot/SIFT 

& $52.3_{0.5}$ & $93.0_{0.4}$ & $92.7_{0.3}$ 
& $7.2_{1.7}$  & $22.1_{1.7}$ & $26.6_{0.4}$ 
& $11.7_{1.4}$ & $65.4_{2.1}$ & $\underline{\mathbf{69.4}}_{3.3}$ 
& $19.4_{0.3}$ & $80.3_{0.3}$ & $78.8_{1.0}$ \\

{} & 10-shot/SIFT 

& $59.8_{0.1}$ & $\underline{\mathbf{93.2}}_{0.2}$ & $\mathbf{92.7}_{0.2}$ 
& $8.7_{2.1}$  & $26.0_{1.6}$ & $29.1_{4.4}$ 
& $15.2_{1.2}$ & $68.1_{0.9}$ & $67.6_{1.3}$ 
& $21.2_{0.1}$ & $69.6_{1.9}$ & $55.6_{7.5}$ \\
\midrule

\multirow{4}{*}{\rotatebox[origin=c]{90}{\shortstack{Llama3}}}   &   0-shot/SFT 

& $3.9_{0.4}$ & $92.8_{0.1}$ & -- 
& $0.0_{0.0}$ & $\mathbf{32.1}_{0.5}$ & -- 
& $0.2_{0.3}$ & $\mathbf{72.2}_{0.7}$ & -- 
& $0.2_{0.0}$ & $\underline{\mathbf{82.8}}_{0.2}$ & -- \\

{} & 1-shot/SIFT 

& $12.1_{0.1}$ & $\mathbf{93.1}_{0.3}$ & $92.6_{0.1}$ 
& $0.8_{0.8}$  & $31.9_{1.1}$ & $32.2_{2.3}$ 
& $2.9_{2.7}$  & $70.5_{1.0}$ & $71.0_{1.0}$ 
& $11.1_{0.5}$ &$82.6_{0.3}$ & $\mathbf{82.5}_{0.3}$ \\

{} & 5-shot/SIFT 

& $20.9_{1.5}$ & $89.4_{3.1}$ & $92.6_{0.2}$ 
& $3.3_{1.5}$  & $26.0_{0.5}$ & $30.4_{1.0}$ 
& $8.2_{0.5}$  & $68.9_{1.8}$ & $69.3_{0.9}$ 
& $16.5_{0.4}$ & $79.5_{0.6}$ & $81.2_{0.2}$ \\

{} & 10-shot/SIFT 

& $20.7_{0.7}$ & $91.9_{0.8}$ & $\underline{\mathbf{93.3}}_{0.1}$ 
& $4.8_{1.7}$  & $29.2_{2.7}$ & $\underline{\mathbf{32.5}}_{1.8}$ 
& $10.1_{2.7}$ & $68.9_{1.6}$ & $\underline{\mathbf{72.6}}_{1.5}$ 
& $16.0_{0.1}$ & $79.8_{2.0}$ & $81.5_{0.5}$ \\
\midrule

\multirow{4}{*}{\rotatebox[origin=c]{90}{\shortstack{Llama3.1 }}}   &   0-shot/SFT 

& $9.9_{0.7}$ & $93.0_{0.2}$ & -- 
& $0.2_{0.3}$ & $\mathbf{30.9}_{1.1}$ & -- 
& $0.0_{0.0}$ & $\underline{\mathbf{73.9}}_{1.0}$ & -- 
& $0.1_{0.0}$ & $\underline{\mathbf{82.4}}_{0.2}$ & -- \\

{} & 1-shot/SIFT 

& $11.2_{0.4}$ & $\mathbf{93.1}_{0.2}$ & $93.3_{0.2}$ 
& $1.8_{0.9}$ & $30.9_{2.1}$ & $32.4_{1.8}$ 
& $2.1_{1.1}$ & $70.5_{0.5}$ & $72.5_{1.1}$ 
& $11.4_{0.1}$ & $82.0_{0.5}$ & $\mathbf{81.9}_{0.3}$ \\

{} & 5-shot/SIFT 

& $19.7_{1.3}$ & $92.1_{0.3}$ & $93.1_{0.2}$ 
& $2.3_{1.6}$  & $26.3_{2.3}$ & $31.6_{2.3}$ 
& $11.2_{2.2}$ & $68.1_{1.3}$ & $72.1_{0.4}$ 
& $18.7_{0.5}$ & $79.2_{0.8}$ & $80.4_{0.2}$ \\

{} & 10-shot/SIFT 

& $24.3_{0.6}$ & $91.5_{0.7}$ & $\underline{\mathbf{93.5}}_{0.2}$ 
& $6.0_{1.0}$  & $30.3_{1.7}$ & $\underline{\mathbf{33.5}}_{2.5}$ 
& $12.1_{1.8}$ & $69.0_{1.1}$ & $\mathbf{72.5}_{0.7}$ 
& $19.1_{0.6}$ & $79.8_{0.5}$ & $81.7_{0.2}$ \\
\midrule

\multirow{4}{*}{\rotatebox[origin=c]{90}{\shortstack{Mistral}}}   &   0-shot/SFT 

& $16.2_{0.6}$ & $93.3_{0.2}$ & -- 
& $0.0_{0.0}$  & $32.7_{2.5}$ & -- 
& $2.1_{0.3}$  & $\mathbf{75.0}_{1.8}$ & -- 
& $0.5_{0.0}$  & $\mathbf{83.9}_{0.2}$ & -- \\

{} & 1-shot/SIFT 

& $37.3_{0.3}$ & $93.3_{0.2}$ & $93.5_{0.3}$ 
& $4.4_{0.5}$  & $34.7_{3.1}$ & $35.7_{2.3}$ 
& $6.2_{1.9}$  & $74.8_{1.4}$ & $\underline{\mathbf{75.6}}_{0.1}$ 
& $19.6_{0.1}$  & $83.9_{0.3}$ & $\underline{\mathbf{83.9}}_{0.1}$ \\

{} & 5-shot/SIFT 

& $52.1_{0.6}$ & $93.2_{0.5}$ & $93.4_{0.2}$ 
& $8.0_{1.5}$  & $32.2_{2.0}$ & $34.9_{2.0}$ 
& $15.3_{2.9}$ & $71.5_{4.0}$ & $75.6_{0.5}$ 
& $23.5_{0.3}$  & $83.1_{0.2}$ & $82.5_{0.3}$ \\

{} & 10-shot/SIFT 

& $54.6_{0.9}$ & $\mathbf{93.4}_{0.1}$ & $\underline{\mathbf{93.5}}_{0.2}$ 
& $11.5_{1.5}$ & $\mathbf{35.3}_{1.3}$ & $\underline{\mathbf{37.7}}_{1.1}$ 
& $21.0_{1.4}$ & $72.8_{0.8}$ & $75.5_{0.5}$ 
& $24.3_{0.1}$ & $74.4_{0.8}$ & $70.4_{6.2}$ \\

\midrule

\multicolumn{2}{c}{LM} & \multicolumn{3}{c}{\hspace{1em}\cancel{\textbf{CM}} \hspace{1em}\textbf{LLM2VEC}} & \multicolumn{3}{c}{\hspace{1em}\cancel{\textbf{CM}} \hspace{1em}\textbf{LLM2VEC}} & \multicolumn{3}{c}{\hspace{1em}\cancel{\textbf{CM}} \hspace{1em}\textbf{LLM2VEC}} & \multicolumn{3}{c}{\hspace{1em}\cancel{\textbf{CM}} \hspace{1em}\textbf{LLM2VEC}} \\
\midrule
\multicolumn{2}{c}{Gemma} 
& \multicolumn{3}{c}{\hspace{-1.5em} $91.3_{0.7}$ \hspace{3em} --} 
& \multicolumn{3}{c}{\hspace{-1em} $0.0_{0.0}$ \hspace{3em} --} 
& \multicolumn{3}{c}{\hspace{-1.5em} $58.8_{4.0}$ \hspace{3em} --} 
& \multicolumn{3}{c}{\hspace{-1.5em} $31.8_{3.5}$ \hspace{3em} --} \\

\multicolumn{2}{c}{Llama2} 

& \multicolumn{3}{c}{$92.0_{0.3}$ \hspace{2em} $59.5_{0.6}$}
& \multicolumn{3}{c}{$0.0_{0.0}$ \hspace{2em} $2.5_{2.1}$} 
& \multicolumn{3}{c}{$51.6_{1.0}$ \hspace{2em} $43.7_{0.8}$} 
& \multicolumn{3}{c}{$31.0_{3.5}$ \hspace{2em} $16.3_{1.9}$} \\

\multicolumn{2}{c}{Llama3} 

& \multicolumn{3}{c}{$92.2_{0.3}$ \hspace{2em} $63.9_{0.5}$} 
& \multicolumn{3}{c}{$6.3_{1.0}$ \hspace{2em} $2.5_{1.3}$} 
& \multicolumn{3}{c}{$69.9_{1.8}$ \hspace{2em} $41.5_{1.9}$} 
& \multicolumn{3}{c}{$28.1_{4.2}$ \hspace{2em} $12.8_{2.8}$} \\

\multicolumn{2}{c}{Llama3.1} 

& \multicolumn{3}{c}{$92.3_{0.5}$ \hspace{2em} $64.4_{0.8}$} 
& \multicolumn{3}{c}{$7.2_{2.1}$ \hspace{2em} $2.2_{1.0}$} 
& \multicolumn{3}{c}{$68.9_{1.6}$ \hspace{2em} $40.4_{0.5}$} 
& \multicolumn{3}{c}{$28.8_{4.1}$ \hspace{2em} $13.2_{2.6}$} \\

\multicolumn{2}{c}{Mistral} 

& \multicolumn{3}{c}{$93.3_{0.2}$ \hspace{2em} $67.3_{0.7}$} 
& \multicolumn{3}{c}{$7.6_{1.4}$ \hspace{2em} $2.7_{0.7}$} 
& \multicolumn{3}{c}{$72.3_{1.1}$ \hspace{2em} $43.9_{1.3}$} 
& \multicolumn{3}{c}{$27.0_{3.9}$ \hspace{2em} $14.5_{3.1}$}\\

\bottomrule
\end{tabular}

}
\caption{Test set micro F1 scores for five \textit{base} decoders on standard SFT and SIFT. We report scores for SRC and MRC strategies and ICL scores with \textit{instruct} decoders on four SL datasets. We also include experiments for the CM removal and LLM2Vec. The decoders are evaluated with the same number of in-context demonstrations (shots) as during fine-tuning. All results are averages of four runs. Standard deviations are shown in subscript. The top two F1 scores per model, dataset, and setup are in bold. Also, the overall best F1 scores between the ICL, SRC, and MRC per model and dataset are underlined.} \label{tab:sift_base_test_results}
\vspace{-0.5em}
\end{table*}
 
\subsection{Main Test Set Results}

\rparagraph{Decoders for Generative Sequence Labeling}
We report the test set results on four SL tasks in \Cref{tab:sift_base_test_results}. However, we do not compare the SRC and MRC strategies with the vanilla CLM, as the vanilla CLM delivered worse validation results. For the same reasons, we omit the SFT and SIFT \textit{instruct} results. However, we do report the ICL results for \textit{instruct} models, as they, expectedly, perform better in ICL than their \textit{base} counterparts. We find, unsurprisingly, that ICL never beats fine-tuning: the trends observed on the validation sets hold on the test sets too. %
Looking at the LM-ing strategies/objectives, we find that, on average, MRC utilizes (more) shots better than SRC. Furthermore, we observe improvements across all datasets except SRL with more shots. For SRL, we obtain the best results with a single demonstration ($1$-shot). We believe that this is because the SRL task comes with $27$ classes, which results in a longer instruction and, directly, a longer context, which negatively impacts performance. 
Notably, $1$-shot SIFT can achieve strong results (cf.~ABAM and SRL tasks), demonstrating the power of even minimal in-context supervision. Consistent with validation performance, we observe lower performance for Llama than for Gemma and Mistral. %

\rparagraph{Decoder-as-Encoder Results}
SIFT outperforms both established decoder-as-encoder baselines: (1) the \textit{CM removal} across all layers in fine-tuning \cite{dukic-snajder-2024-looking} and (2) fine-tuning of the popular \textit{LLM2Vec} approach \cite{behnamghaderllm2vec}. CM removal performs on a par with our best-performing SIFT models on NER, but falls well behind SIFT on other datasets, suggesting that perhaps a more nuanced, task-specific unmasking configurations are needed. LLM2Vec falls short of CM removal in performance, suggesting that generating sequence embeddings---a key objective in LLM2Vec's conversion of a decoder into an encoder in a task-agnostic manner---does not necessarily translate into informative token representations, which are critical for SL.

\subsection{Instruction Ablation Results}

\begin{figure}
    \includegraphics[width=\columnwidth]{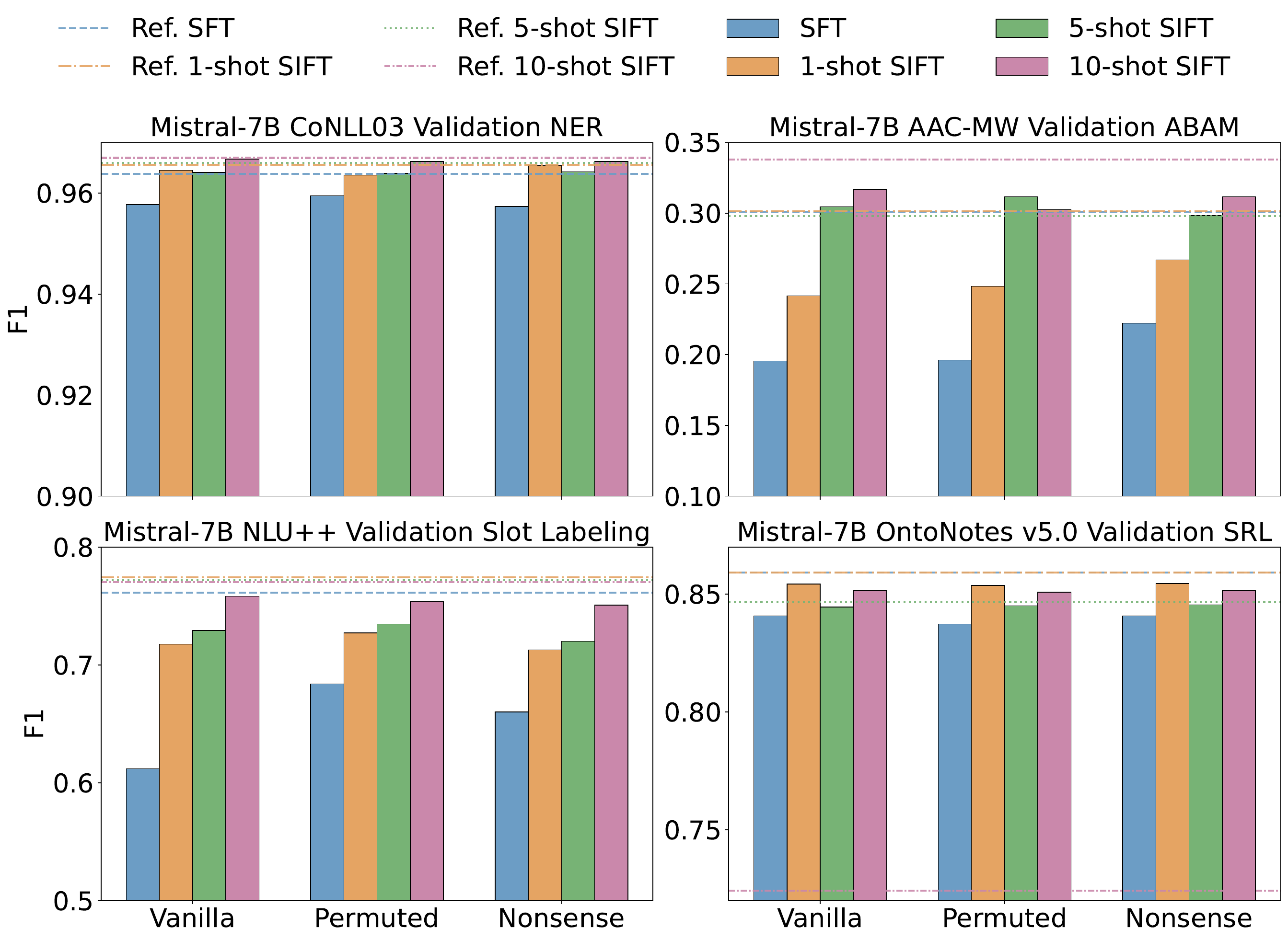}
    \caption{
        Instruction variants applied on the validation set at inference time for Mistral-7B (\textit{base}) models trained with MRC and without instructions. The reference (Ref.) lines show the results for the same models trained with instructions. All results are averages of four runs.
    }
    \label{fig:instruction_effect_mistral}
\end{figure}

We report the results for Mistral-7B (see \Cref{fig:instruction_effect_mistral}) and Gemma-7B (see \Cref{fig:instruction_effect_gemma} in \Cref{sec:complementary_res}) trained with MRC---our generally best-performing SIFT strategy. We observe that including the instruction has a significantly negative impact on SFT performance, but to a lesser extent in SIFT (for models trained without the instruction). The SIFT performance is expectedly worse compared to having the instruction included in both training and inference, but the gap narrows with more in-context demonstrations. This suggests that at inference time, a SIFT model (trained without instruction) benefits more from informative in-context demonstrations than from the instruction. 
Interestingly, for SRL, SIFT models with more shots (see $10$-shot) trained without the instruction even perform better than their counterparts trained with the instruction: we believe that this is due to the drastic context length increase brought about by the instruction---due to the SRL's large number of classes (27)---which reduces the model's ability to effectively ``consume'' the entire context including the provided in-context demonstrations.

\section{Related Work}

\rparagraph{Learning to Learn In-Context} 
The field has explored SIFT method under the \textit{learning to learn in context} umbrella, explicitly improving ICL \citep{zhuang2024vector,li2025large}. The method was first introduced under the pseudonyms \textit{meta ICL} \citep{min-etal-2022-metaicl} and \textit{in-context tuning} \citep{chen-etal-2022-meta}. Few-shot learning was also explored and compared with ICL on SL tasks, such as slot labeling and NER \citep{chen-etal-2023-learning, razumovskaia2024analyzing}. However, these works only consider \textit{vanilla} SIFT (i.e., fine-tune on all prompt tokens). Other works propose fine-tuning the model on the response(s) only (i.e., not on the instruction part of the prompt), but do not conduct few-shot learning \citep{an2024response,hewitt2024instruction}. In contrast, we propose novel, response adaptation strategies in SIFT framework for SL (SRC and MRC, see \S\ref{sec:sift}) for utilizing many demonstrations during fine-tuning and inference.

\rparagraph{In-Context Learning vs.~Fine-Tuning} 
ICL has been compared to fine-tuning in the literature numerous times \citep{duan-etal-2024-exploring, mosbach-etal-2023-shot, yin-etal-2024-deeper}. The verdict is that once the model is given enough training examples, fine-tuning beats ICL \citep{bertsch-etal-2025-context}. The same authors find that LLMs typically improve performance when presented with more demonstrations, especially for datasets with large label spaces. In contrast, \citet{li2025longcontext} report that long-context ICL achieves subpar performance with the increase of the number of demonstrations, highlighting once more that LLMs struggle to utilize the long contexts \citep{10.1162/tacl_a_00638}. We investigated whether and to what extent this holds for SL. 

\rparagraph{Models for Sequence Labeling} 
The first SL models used manually designed features \citep{he2020survey} that were fed into statistical models \citep{tjong-kim-sang-buchholz-2000-introduction, ahn-2006-stages} such as support vector machines, maximum entropy Markov models, or conditional random fields \citep{lafferty2001conditional}. Focus later shifted to recurrent neural networks, namely BiLSTMs \citep{akbik-etal-2018-contextual}. Following the Transformer era \citep{vaswani2017attention}, SL was addressed with encoder-only models \citep{devlin-etal-2019-bert, fei-etal-2021-better-new}, pre-trained with a bidirectional language modeling objective beneficial for most NLU tasks. More recently, SL has also been performed autoregressively—by generating spans and their labels in natural language \citep{wang2022instructionner, wang2023instructuie}—where encoder-decoder models have proven useful.
With the rise of pre-training, decoder-only models became prominent. Unlike encoders, decoders are constrained by causal masking, which prevents them from attending to future tokens, thereby limiting their effectiveness on tasks that require bidirectionality. \citet{dukic-snajder-2024-looking} mitigate this limitation by removing the CM in a subset of layers, allowing the decoder to function as an encoder and unlocking its potential for SL. Similarly, \citet{behnamghaderllm2vec} propose an LLM2Vec approach to transform decoders into encoders using specialized training objectives.
However, the autoregressive potential of decoders for SL has been less explored. We investigate this potential by combining in-context adaptation, ICL, and constrained decoding.

\rparagraph{Loss Function Modification} 
A typical pre-trained neural language model is adapted to the target task by modifying the loss function, which usually involves adding a regularization term to prevent overfitting and avoid catastrophic forgetting \citep{wiese-etal-2017-neural}. Some methods indirectly affect parameter updates through the loss function. For example, embeddings of the auxiliary task labels can be used to guide the parameter updates through the loss function \citep{dukic-etal-2024-leveraging}. Furthermore, some tokens can be included in the loss calculation, and others can be excluded. \citet{huerta-enochian-ko-2024-instruction} explore the impact of excluding tokens from loss during instruction tuning of decoder models. However, a more practical approach for decoders would be to compute the loss only on expected responses, excluding preceding tokens in the prompt \citep{an2024response}. This adjustment penalizes the model only for incorrectly generated response tokens, aligning with our proposed SRC strategy without demonstrations in the context. However, including multiple demonstrations in the context and performing CLM only over response tokens for SL has not yet been explored in the community.
\section{Conclusion}

Since the advent of LLMs, the question of how to best utilize them for sequence labeling (SL) has remained open. Our framework demonstrates that the key ingredients for maximizing LLM performance on SL are (1) using many demonstrations during fine-tuning (SIFT) and inference (ICL), (2) tuning with multi-response completion (MRC), (3) predicting with constrained generation, and (4) relying on the original instruction. 

While this work used randomly sampled demonstrations, the framework's effectiveness could likely be boosted by integrating methods that select more informative examples. Such sampling could be incorporated during both fine-tuning and inference, as explored in prior work \citep{rubin-etal-2022-learning,ye2023compositional}. Moreover, to address decoders' struggles with long contexts, SIFT methods could be enhanced by integrating approaches for disentangling latent shifts \citep{10.5555/3692070.3693379, jukic2024disentangling}. 
\bibliography{custom,anthology}

\begin{thebibliography}{67}
\expandafter\ifx\csname natexlab\endcsname\relax\def\natexlab#1{#1}\fi

\bibitem[{Agarwal et~al.(2024)Agarwal, Singh, Zhang, Bohnet, Rosias, Chan,
  Zhang, Anand, Abbas, Nova et~al.}]{agarwal2024many}
Rishabh Agarwal, Avi Singh, Lei Zhang, Bernd Bohnet, Luis Rosias, Stephanie
  Chan, Biao Zhang, Ankesh Anand, Zaheer Abbas, Azade Nova, et~al. 2024.
\newblock Many-shot in-context learning.
\newblock \emph{Advances in Neural Information Processing Systems},
  37:76930--76966.

\bibitem[{Ahn(2006)}]{ahn-2006-stages}
David Ahn. 2006.
\newblock \href {https://aclanthology.org/W06-0901/} {The stages of event
  extraction}.
\newblock In \emph{Proceedings of the Workshop on Annotating and Reasoning
  about Time and Events}, pages 1--8, Sydney, Australia. Association for
  Computational Linguistics.

\bibitem[{Akbik et~al.(2018)Akbik, Blythe, and
  Vollgraf}]{akbik-etal-2018-contextual}
Alan Akbik, Duncan Blythe, and Roland Vollgraf. 2018.
\newblock \href {https://aclanthology.org/C18-1139/} {Contextual string
  embeddings for sequence labeling}.
\newblock In \emph{Proceedings of the 27th International Conference on
  Computational Linguistics}, pages 1638--1649, Santa Fe, New Mexico, USA.
  Association for Computational Linguistics.

\bibitem[{An and Kim(2024)}]{an2024response}
Seokhyun An and Hyounghun Kim. 2024.
\newblock Response tuning: Aligning large language models without instruction.
\newblock \emph{arXiv preprint arXiv:2410.02465}.

\bibitem[{BehnamGhader et~al.(2024)BehnamGhader, Adlakha, Mosbach, Bahdanau,
  Chapados, and Reddy}]{behnamghaderllm2vec}
Parishad BehnamGhader, Vaibhav Adlakha, Marius Mosbach, Dzmitry Bahdanau,
  Nicolas Chapados, and Siva Reddy. 2024.
\newblock Llm2vec: Large language models are secretly powerful text encoders.
\newblock In \emph{First Conference on Language Modeling}.

\bibitem[{Bertsch et~al.(2025)Bertsch, Ivgi, Xiao, Alon, Berant, Gormley, and
  Neubig}]{bertsch-etal-2025-context}
Amanda Bertsch, Maor Ivgi, Emily Xiao, Uri Alon, Jonathan Berant, Matthew~R.
  Gormley, and Graham Neubig. 2025.
\newblock \href {https://doi.org/10.18653/v1/2025.naacl-long.605} {In-context
  learning with long-context models: An in-depth exploration}.
\newblock In \emph{Proceedings of the 2025 Conference of the Nations of the
  Americas Chapter of the Association for Computational Linguistics: Human
  Language Technologies (Volume 1: Long Papers)}, pages 12119--12149,
  Albuquerque, New Mexico. Association for Computational Linguistics.

\bibitem[{Brown et~al.(2020)Brown, Mann, Ryder, Subbiah, Kaplan, Dhariwal,
  Neelakantan, Shyam, Sastry, Askell et~al.}]{brown2020language}
Tom Brown, Benjamin Mann, Nick Ryder, Melanie Subbiah, Jared~D Kaplan, Prafulla
  Dhariwal, Arvind Neelakantan, Pranav Shyam, Girish Sastry, Amanda Askell,
  et~al. 2020.
\newblock Language models are few-shot learners.
\newblock \emph{Advances in neural information processing systems},
  33:1877--1901.

\bibitem[{Casanueva et~al.(2022)Casanueva, Vuli{\'c}, Spithourakis, and
  Budzianowski}]{casanueva-etal-2022-nlu}
Inigo Casanueva, Ivan Vuli{\'c}, Georgios Spithourakis, and Pawe{\l}
  Budzianowski. 2022.
\newblock \href {https://doi.org/10.18653/v1/2022.findings-naacl.154} {{NLU}++:
  A multi-label, slot-rich, generalisable dataset for natural language
  understanding in task-oriented dialogue}.
\newblock In \emph{Findings of the Association for Computational Linguistics:
  NAACL 2022}, pages 1998--2013, Seattle, United States. Association for
  Computational Linguistics.

\bibitem[{Chen et~al.(2023)Chen, Lu, Lin, Lou, Jia, Dai, Wu, Cao, Han, and
  Sun}]{chen-etal-2023-learning}
Jiawei Chen, Yaojie Lu, Hongyu Lin, Jie Lou, Wei Jia, Dai Dai, Hua Wu, Boxi
  Cao, Xianpei Han, and Le~Sun. 2023.
\newblock \href {https://doi.org/10.18653/v1/2023.acl-long.764} {Learning
  in-context learning for named entity recognition}.
\newblock In \emph{Proceedings of the 61st Annual Meeting of the Association
  for Computational Linguistics (Volume 1: Long Papers)}, pages 13661--13675,
  Toronto, Canada. Association for Computational Linguistics.

\bibitem[{Chen et~al.(2022)Chen, Zhong, Zha, Karypis, and
  He}]{chen-etal-2022-meta}
Yanda Chen, Ruiqi Zhong, Sheng Zha, George Karypis, and He~He. 2022.
\newblock \href {https://doi.org/10.18653/v1/2022.acl-long.53} {Meta-learning
  via language model in-context tuning}.
\newblock In \emph{Proceedings of the 60th Annual Meeting of the Association
  for Computational Linguistics (Volume 1: Long Papers)}, pages 719--730,
  Dublin, Ireland. Association for Computational Linguistics.

\bibitem[{Dettmers et~al.(2023)Dettmers, Pagnoni, Holtzman, and
  Zettlemoyer}]{dettmers2023qlora}
Tim Dettmers, Artidoro Pagnoni, Ari Holtzman, and Luke Zettlemoyer. 2023.
\newblock \href {https://openreview.net/forum?id=OUIFPHEgJU} {{QL}o{RA}:
  Efficient finetuning of quantized {LLM}s}.
\newblock In \emph{Thirty-seventh Conference on Neural Information Processing
  Systems}.

\bibitem[{Devlin et~al.(2019)Devlin, Chang, Lee, and
  Toutanova}]{devlin-etal-2019-bert}
Jacob Devlin, Ming-Wei Chang, Kenton Lee, and Kristina Toutanova. 2019.
\newblock \href {https://doi.org/10.18653/v1/N19-1423} {{BERT}: Pre-training of
  deep bidirectional transformers for language understanding}.
\newblock In \emph{Proceedings of the 2019 Conference of the North {A}merican
  Chapter of the Association for Computational Linguistics: Human Language
  Technologies, Volume 1 (Long and Short Papers)}, pages 4171--4186,
  Minneapolis, Minnesota. Association for Computational Linguistics.

\bibitem[{Duan et~al.(2024)Duan, Tang, Yang, Abbasi, and
  Tam}]{duan-etal-2024-exploring}
Hanyu Duan, Yixuan Tang, Yi~Yang, Ahmed Abbasi, and Kar~Yan Tam. 2024.
\newblock \href {https://doi.org/10.18653/v1/2024.findings-emnlp.182}
  {Exploring the relationship between in-context learning and instruction
  tuning}.
\newblock In \emph{Findings of the Association for Computational Linguistics:
  EMNLP 2024}, pages 3197--3210, Miami, Florida, USA. Association for
  Computational Linguistics.

\bibitem[{Duki{\'c} et~al.(2024{\natexlab{a}})Duki{\'c}, Do{\v{s}}ilovi{\'c},
  Plu{\v{s}}{\v{c}}ec, and {\v{S}}najder}]{dukic2024closed}
David Duki{\'c}, Filip~Karlo Do{\v{s}}ilovi{\'c}, Domagoj Plu{\v{s}}{\v{c}}ec,
  and Jan {\v{S}}najder. 2024{\natexlab{a}}.
\newblock Closed-domain event extraction for hard news event monitoring: {A}
  systematic study.
\newblock \emph{PeerJ Computer Science}, 10:e2355.

\bibitem[{Duki{\'c} et~al.(2024{\natexlab{b}})Duki{\'c}, Gashteovski,
  Glava{\v{s}}, and Snajder}]{dukic-etal-2024-leveraging}
David Duki{\'c}, Kiril Gashteovski, Goran Glava{\v{s}}, and Jan Snajder.
  2024{\natexlab{b}}.
\newblock \href {https://aclanthology.org/2024.findings-eacl.80/} {Leveraging
  open information extraction for more robust domain transfer of event trigger
  detection}.
\newblock In \emph{Findings of the Association for Computational Linguistics:
  EACL 2024}, pages 1197--1213, St. Julian{'}s, Malta. Association for
  Computational Linguistics.

\bibitem[{Duki{\'c} et~al.(2024{\natexlab{c}})Duki{\'c}, Petri{\v{c}}evi{\'c},
  {\'C}urkovi{\'c}, and {\v{S}}najder}]{dukic2024takelab}
David Duki{\'c}, Marin Petri{\v{c}}evi{\'c}, Sven {\'C}urkovi{\'c}, and Jan
  {\v{S}}najder. 2024{\natexlab{c}}.
\newblock Takelab {Retriever}: {AI}-driven search engine for articles from
  {Croatian} news outlets.
\newblock \emph{arXiv preprint arXiv:2411.19718}.

\bibitem[{Duki{\'c} and {\v{S}}najder(2024)}]{dukic-snajder-2024-looking}
David Duki{\'c} and Jan {\v{S}}najder. 2024.
\newblock \href {https://doi.org/10.18653/v1/2024.findings-acl.843} {Looking
  right is sometimes right: Investigating the capabilities of decoder-only
  {LLM}s for sequence labeling}.
\newblock In \emph{Findings of the Association for Computational Linguistics:
  ACL 2024}, pages 14168--14181, Bangkok, Thailand. Association for
  Computational Linguistics.

\bibitem[{Fei et~al.(2021)Fei, Wu, Ren, Li, and Ji}]{fei-etal-2021-better-new}
Hao Fei, Shengqiong Wu, Yafeng Ren, Fei Li, and Donghong Ji. 2021.
\newblock \href {https://doi.org/10.18653/v1/2021.findings-acl.49} {Better
  combine them together! {Integrating} syntactic constituency and dependency
  representations for semantic role labeling}.
\newblock In \emph{Findings of the Association for Computational Linguistics:
  ACL-IJCNLP 2021}, pages 549--559, Online. Association for Computational
  Linguistics.

\bibitem[{Grattafiori et~al.(2024)Grattafiori, Dubey, Jauhri, Pandey, Kadian,
  Al-Dahle, Letman, Mathur, Schelten, Vaughan, Yang, Fan, Goyal, Hartshorn,
  Yang, Mitra, Sravankumar, Korenev, Hinsvark, Rao, Zhang, Rodriguez,
  Gregerson, Spataru, Roziere, Biron, Tang, Chern, Caucheteux, Nayak, Bi,
  Marra, McConnell, Keller, Touret, Wu, Wong, Ferrer, Nikolaidis, Allonsius,
  Song, Pintz, Livshits, Wyatt, Esiobu, Choudhary, Mahajan, Garcia-Olano,
  Perino, Hupkes, Lakomkin, AlBadawy, Lobanova, Dinan, Smith, Radenovic,
  Guzmán, Zhang, Synnaeve, Lee, Anderson, Thattai, Nail, Mialon, Pang,
  Cucurell, Nguyen, Korevaar, Xu, Touvron, Zarov, Ibarra, Kloumann, Misra,
  Evtimov, Zhang, Copet, Lee, Geffert, Vranes, Park, Mahadeokar, Shah, van~der
  Linde, Billock, Hong, Lee, Fu, Chi, Huang, Liu, Wang, Yu, Bitton, Spisak,
  Park, Rocca, Johnstun, Saxe, Jia, Alwala, Prasad, Upasani, Plawiak, Li,
  Heafield, Stone, El-Arini, Iyer, Malik, Chiu, Bhalla, Lakhotia,
  Rantala-Yeary, van~der Maaten, Chen, Tan, Jenkins, Martin, Madaan, Malo,
  Blecher, Landzaat, de~Oliveira, Muzzi, Pasupuleti, Singh, Paluri, Kardas,
  Tsimpoukelli, Oldham, Rita, Pavlova, Kambadur, Lewis, Si, Singh, Hassan,
  Goyal, Torabi, Bashlykov, Bogoychev, Chatterji, Zhang, Duchenne, Çelebi,
  Alrassy, Zhang, Li, Vasic, Weng, Bhargava, Dubal, Krishnan, Koura, Xu, He,
  Dong, Srinivasan, Ganapathy, Calderer, Cabral, Stojnic, Raileanu, Maheswari,
  Girdhar, Patel, Sauvestre, Polidoro, Sumbaly, Taylor, Silva, Hou, Wang,
  Hosseini, Chennabasappa, Singh, Bell, Kim, Edunov, Nie, Narang, Raparthy,
  Shen, Wan, Bhosale, Zhang, Vandenhende, Batra, Whitman, Sootla, Collot,
  Gururangan, Borodinsky, Herman, Fowler, Sheasha, Georgiou, Scialom,
  Speckbacher, Mihaylov, Xiao, Karn, Goswami, Gupta, Ramanathan, Kerkez,
  Gonguet, Do, Vogeti, Albiero, Petrovic, Chu, Xiong, Fu, Meers, Martinet,
  Wang, Wang, Tan, Xia, Xie, Jia, Wang, Goldschlag, Gaur, Babaei, Wen, Song,
  Zhang, Li, Mao, Coudert, Yan, Chen, Papakipos, Singh, Srivastava, Jain,
  Kelsey, Shajnfeld, Gangidi, Victoria, Goldstand, Menon, Sharma, Boesenberg,
  Baevski, Feinstein, Kallet, Sangani, Teo, Yunus, Lupu, Alvarado, Caples, Gu,
  Ho, Poulton, Ryan, Ramchandani, Dong, Franco, Goyal, Saraf, Chowdhury,
  Gabriel, Bharambe, Eisenman, Yazdan, James, Maurer, Leonhardi, Huang, Loyd,
  Paola, Paranjape, Liu, Wu, Ni, Hancock, Wasti, Spence, Stojkovic, Gamido,
  Montalvo, Parker, Burton, Mejia, Liu, Wang, Kim, Zhou, Hu, Chu, Cai, Tindal,
  Feichtenhofer, Gao, Civin, Beaty, Kreymer, Li, Adkins, Xu, Testuggine, David,
  Parikh, Liskovich, Foss, Wang, Le, Holland, Dowling, Jamil, Montgomery,
  Presani, Hahn, Wood, Le, Brinkman, Arcaute, Dunbar, Smothers, Sun, Kreuk,
  Tian, Kokkinos, Ozgenel, Caggioni, Kanayet, Seide, Florez, Schwarz, Badeer,
  Swee, Halpern, Herman, Sizov, Guangyi, Zhang, Lakshminarayanan, Inan,
  Shojanazeri, Zou, Wang, Zha, Habeeb, Rudolph, Suk, Aspegren, Goldman, Zhan,
  Damlaj, Molybog, Tufanov, Leontiadis, Veliche, Gat, Weissman, Geboski, Kohli,
  Lam, Asher, Gaya, Marcus, Tang, Chan, Zhen, Reizenstein, Teboul, Zhong, Jin,
  Yang, Cummings, Carvill, Shepard, McPhie, Torres, Ginsburg, Wang, Wu, U,
  Saxena, Khandelwal, Zand, Matosich, Veeraraghavan, Michelena, Li, Jagadeesh,
  Huang, Chawla, Huang, Chen, Garg, A, Silva, Bell, Zhang, Guo, Yu, Moshkovich,
  Wehrstedt, Khabsa, Avalani, Bhatt, Mankus, Hasson, Lennie, Reso, Groshev,
  Naumov, Lathi, Keneally, Liu, Seltzer, Valko, Restrepo, Patel, Vyatskov,
  Samvelyan, Clark, Macey, Wang, Hermoso, Metanat, Rastegari, Bansal,
  Santhanam, Parks, White, Bawa, Singhal, Egebo, Usunier, Mehta, Laptev, Dong,
  Cheng, Chernoguz, Hart, Salpekar, Kalinli, Kent, Parekh, Saab, Balaji,
  Rittner, Bontrager, Roux, Dollar, Zvyagina, Ratanchandani, Yuvraj, Liang,
  Alao, Rodriguez, Ayub, Murthy, Nayani, Mitra, Parthasarathy, Li, Hogan,
  Battey, Wang, Howes, Rinott, Mehta, Siby, Bondu, Datta, Chugh, Hunt, Dhillon,
  Sidorov, Pan, Mahajan, Verma, Yamamoto, Ramaswamy, Lindsay, Lindsay, Feng,
  Lin, Zha, Patil, Shankar, Zhang, Zhang, Wang, Agarwal, Sajuyigbe, Chintala,
  Max, Chen, Kehoe, Satterfield, Govindaprasad, Gupta, Deng, Cho, Virk,
  Subramanian, Choudhury, Goldman, Remez, Glaser, Best, Koehler, Robinson, Li,
  Zhang, Matthews, Chou, Shaked, Vontimitta, Ajayi, Montanez, Mohan, Kumar,
  Mangla, Ionescu, Poenaru, Mihailescu, Ivanov, Li, Wang, Jiang, Bouaziz,
  Constable, Tang, Wu, Wang, Wu, Gao, Kleinman, Chen, Hu, Jia, Qi, Li, Zhang,
  Zhang, Adi, Nam, Yu, Wang, Zhao, Hao, Qian, Li, He, Rait, DeVito, Rosnbrick,
  Wen, Yang, Zhao, and Ma}]{dubey2024llama}
Aaron Grattafiori, Abhimanyu Dubey, Abhinav Jauhri, Abhinav Pandey, Abhishek
  Kadian, Ahmad Al-Dahle, Aiesha Letman, Akhil Mathur, Alan Schelten, Alex
  Vaughan, Amy Yang, Angela Fan, Anirudh Goyal, Anthony Hartshorn, Aobo Yang,
  Archi Mitra, Archie Sravankumar, Artem Korenev, Arthur Hinsvark, Arun Rao,
  Aston Zhang, Aurelien Rodriguez, Austen Gregerson, Ava Spataru, Baptiste
  Roziere, Bethany Biron, Binh Tang, Bobbie Chern, Charlotte Caucheteux, Chaya
  Nayak, Chloe Bi, Chris Marra, Chris McConnell, Christian Keller, Christophe
  Touret, Chunyang Wu, Corinne Wong, Cristian~Canton Ferrer, Cyrus Nikolaidis,
  Damien Allonsius, Daniel Song, Danielle Pintz, Danny Livshits, Danny Wyatt,
  David Esiobu, Dhruv Choudhary, Dhruv Mahajan, Diego Garcia-Olano, Diego
  Perino, Dieuwke Hupkes, Egor Lakomkin, Ehab AlBadawy, Elina Lobanova, Emily
  Dinan, Eric~Michael Smith, Filip Radenovic, Francisco Guzmán, Frank Zhang,
  Gabriel Synnaeve, Gabrielle Lee, Georgia~Lewis Anderson, Govind Thattai,
  Graeme Nail, Gregoire Mialon, Guan Pang, Guillem Cucurell, Hailey Nguyen,
  Hannah Korevaar, Hu~Xu, Hugo Touvron, Iliyan Zarov, Imanol~Arrieta Ibarra,
  Isabel Kloumann, Ishan Misra, Ivan Evtimov, Jack Zhang, Jade Copet, Jaewon
  Lee, Jan Geffert, Jana Vranes, Jason Park, Jay Mahadeokar, Jeet Shah, Jelmer
  van~der Linde, Jennifer Billock, Jenny Hong, Jenya Lee, Jeremy Fu, Jianfeng
  Chi, Jianyu Huang, Jiawen Liu, Jie Wang, Jiecao Yu, Joanna Bitton, Joe
  Spisak, Jongsoo Park, Joseph Rocca, Joshua Johnstun, Joshua Saxe, Junteng
  Jia, Kalyan~Vasuden Alwala, Karthik Prasad, Kartikeya Upasani, Kate Plawiak,
  Ke~Li, Kenneth Heafield, Kevin Stone, Khalid El-Arini, Krithika Iyer, Kshitiz
  Malik, Kuenley Chiu, Kunal Bhalla, Kushal Lakhotia, Lauren Rantala-Yeary,
  Laurens van~der Maaten, Lawrence Chen, Liang Tan, Liz Jenkins, Louis Martin,
  Lovish Madaan, Lubo Malo, Lukas Blecher, Lukas Landzaat, Luke de~Oliveira,
  Madeline Muzzi, Mahesh Pasupuleti, Mannat Singh, Manohar Paluri, Marcin
  Kardas, Maria Tsimpoukelli, Mathew Oldham, Mathieu Rita, Maya Pavlova,
  Melanie Kambadur, Mike Lewis, Min Si, Mitesh~Kumar Singh, Mona Hassan, Naman
  Goyal, Narjes Torabi, Nikolay Bashlykov, Nikolay Bogoychev, Niladri
  Chatterji, Ning Zhang, Olivier Duchenne, Onur Çelebi, Patrick Alrassy,
  Pengchuan Zhang, Pengwei Li, Petar Vasic, Peter Weng, Prajjwal Bhargava,
  Pratik Dubal, Praveen Krishnan, Punit~Singh Koura, Puxin Xu, Qing He,
  Qingxiao Dong, Ragavan Srinivasan, Raj Ganapathy, Ramon Calderer,
  Ricardo~Silveira Cabral, Robert Stojnic, Roberta Raileanu, Rohan Maheswari,
  Rohit Girdhar, Rohit Patel, Romain Sauvestre, Ronnie Polidoro, Roshan
  Sumbaly, Ross Taylor, Ruan Silva, Rui Hou, Rui Wang, Saghar Hosseini, Sahana
  Chennabasappa, Sanjay Singh, Sean Bell, Seohyun~Sonia Kim, Sergey Edunov,
  Shaoliang Nie, Sharan Narang, Sharath Raparthy, Sheng Shen, Shengye Wan,
  Shruti Bhosale, Shun Zhang, Simon Vandenhende, Soumya Batra, Spencer Whitman,
  Sten Sootla, Stephane Collot, Suchin Gururangan, Sydney Borodinsky, Tamar
  Herman, Tara Fowler, Tarek Sheasha, Thomas Georgiou, Thomas Scialom, Tobias
  Speckbacher, Todor Mihaylov, Tong Xiao, Ujjwal Karn, Vedanuj Goswami, Vibhor
  Gupta, Vignesh Ramanathan, Viktor Kerkez, Vincent Gonguet, Virginie Do, Vish
  Vogeti, Vítor Albiero, Vladan Petrovic, Weiwei Chu, Wenhan Xiong, Wenyin Fu,
  Whitney Meers, Xavier Martinet, Xiaodong Wang, Xiaofang Wang, Xiaoqing~Ellen
  Tan, Xide Xia, Xinfeng Xie, Xuchao Jia, Xuewei Wang, Yaelle Goldschlag,
  Yashesh Gaur, Yasmine Babaei, Yi~Wen, Yiwen Song, Yuchen Zhang, Yue Li,
  Yuning Mao, Zacharie~Delpierre Coudert, Zheng Yan, Zhengxing Chen, Zoe
  Papakipos, Aaditya Singh, Aayushi Srivastava, Abha Jain, Adam Kelsey, Adam
  Shajnfeld, Adithya Gangidi, Adolfo Victoria, Ahuva Goldstand, Ajay Menon,
  Ajay Sharma, Alex Boesenberg, Alexei Baevski, Allie Feinstein, Amanda Kallet,
  Amit Sangani, Amos Teo, Anam Yunus, Andrei Lupu, Andres Alvarado, Andrew
  Caples, Andrew Gu, Andrew Ho, Andrew Poulton, Andrew Ryan, Ankit Ramchandani,
  Annie Dong, Annie Franco, Anuj Goyal, Aparajita Saraf, Arkabandhu Chowdhury,
  Ashley Gabriel, Ashwin Bharambe, Assaf Eisenman, Azadeh Yazdan, Beau James,
  Ben Maurer, Benjamin Leonhardi, Bernie Huang, Beth Loyd, Beto~De Paola,
  Bhargavi Paranjape, Bing Liu, Bo~Wu, Boyu Ni, Braden Hancock, Bram Wasti,
  Brandon Spence, Brani Stojkovic, Brian Gamido, Britt Montalvo, Carl Parker,
  Carly Burton, Catalina Mejia, Ce~Liu, Changhan Wang, Changkyu Kim, Chao Zhou,
  Chester Hu, Ching-Hsiang Chu, Chris Cai, Chris Tindal, Christoph
  Feichtenhofer, Cynthia Gao, Damon Civin, Dana Beaty, Daniel Kreymer, Daniel
  Li, David Adkins, David Xu, Davide Testuggine, Delia David, Devi Parikh,
  Diana Liskovich, Didem Foss, Dingkang Wang, Duc Le, Dustin Holland, Edward
  Dowling, Eissa Jamil, Elaine Montgomery, Eleonora Presani, Emily Hahn, Emily
  Wood, Eric-Tuan Le, Erik Brinkman, Esteban Arcaute, Evan Dunbar, Evan
  Smothers, Fei Sun, Felix Kreuk, Feng Tian, Filippos Kokkinos, Firat Ozgenel,
  Francesco Caggioni, Frank Kanayet, Frank Seide, Gabriela~Medina Florez,
  Gabriella Schwarz, Gada Badeer, Georgia Swee, Gil Halpern, Grant Herman,
  Grigory Sizov, Guangyi, Zhang, Guna Lakshminarayanan, Hakan Inan, Hamid
  Shojanazeri, Han Zou, Hannah Wang, Hanwen Zha, Haroun Habeeb, Harrison
  Rudolph, Helen Suk, Henry Aspegren, Hunter Goldman, Hongyuan Zhan, Ibrahim
  Damlaj, Igor Molybog, Igor Tufanov, Ilias Leontiadis, Irina-Elena Veliche,
  Itai Gat, Jake Weissman, James Geboski, James Kohli, Janice Lam, Japhet
  Asher, Jean-Baptiste Gaya, Jeff Marcus, Jeff Tang, Jennifer Chan, Jenny Zhen,
  Jeremy Reizenstein, Jeremy Teboul, Jessica Zhong, Jian Jin, Jingyi Yang, Joe
  Cummings, Jon Carvill, Jon Shepard, Jonathan McPhie, Jonathan Torres, Josh
  Ginsburg, Junjie Wang, Kai Wu, Kam~Hou U, Karan Saxena, Kartikay Khandelwal,
  Katayoun Zand, Kathy Matosich, Kaushik Veeraraghavan, Kelly Michelena, Keqian
  Li, Kiran Jagadeesh, Kun Huang, Kunal Chawla, Kyle Huang, Lailin Chen,
  Lakshya Garg, Lavender A, Leandro Silva, Lee Bell, Lei Zhang, Liangpeng Guo,
  Licheng Yu, Liron Moshkovich, Luca Wehrstedt, Madian Khabsa, Manav Avalani,
  Manish Bhatt, Martynas Mankus, Matan Hasson, Matthew Lennie, Matthias Reso,
  Maxim Groshev, Maxim Naumov, Maya Lathi, Meghan Keneally, Miao Liu,
  Michael~L. Seltzer, Michal Valko, Michelle Restrepo, Mihir Patel, Mik
  Vyatskov, Mikayel Samvelyan, Mike Clark, Mike Macey, Mike Wang, Miquel~Jubert
  Hermoso, Mo~Metanat, Mohammad Rastegari, Munish Bansal, Nandhini Santhanam,
  Natascha Parks, Natasha White, Navyata Bawa, Nayan Singhal, Nick Egebo,
  Nicolas Usunier, Nikhil Mehta, Nikolay~Pavlovich Laptev, Ning Dong, Norman
  Cheng, Oleg Chernoguz, Olivia Hart, Omkar Salpekar, Ozlem Kalinli, Parkin
  Kent, Parth Parekh, Paul Saab, Pavan Balaji, Pedro Rittner, Philip Bontrager,
  Pierre Roux, Piotr Dollar, Polina Zvyagina, Prashant Ratanchandani, Pritish
  Yuvraj, Qian Liang, Rachad Alao, Rachel Rodriguez, Rafi Ayub, Raghotham
  Murthy, Raghu Nayani, Rahul Mitra, Rangaprabhu Parthasarathy, Raymond Li,
  Rebekkah Hogan, Robin Battey, Rocky Wang, Russ Howes, Ruty Rinott, Sachin
  Mehta, Sachin Siby, Sai~Jayesh Bondu, Samyak Datta, Sara Chugh, Sara Hunt,
  Sargun Dhillon, Sasha Sidorov, Satadru Pan, Saurabh Mahajan, Saurabh Verma,
  Seiji Yamamoto, Sharadh Ramaswamy, Shaun Lindsay, Shaun Lindsay, Sheng Feng,
  Shenghao Lin, Shengxin~Cindy Zha, Shishir Patil, Shiva Shankar, Shuqiang
  Zhang, Shuqiang Zhang, Sinong Wang, Sneha Agarwal, Soji Sajuyigbe, Soumith
  Chintala, Stephanie Max, Stephen Chen, Steve Kehoe, Steve Satterfield,
  Sudarshan Govindaprasad, Sumit Gupta, Summer Deng, Sungmin Cho, Sunny Virk,
  Suraj Subramanian, Sy~Choudhury, Sydney Goldman, Tal Remez, Tamar Glaser,
  Tamara Best, Thilo Koehler, Thomas Robinson, Tianhe Li, Tianjun Zhang, Tim
  Matthews, Timothy Chou, Tzook Shaked, Varun Vontimitta, Victoria Ajayi,
  Victoria Montanez, Vijai Mohan, Vinay~Satish Kumar, Vishal Mangla, Vlad
  Ionescu, Vlad Poenaru, Vlad~Tiberiu Mihailescu, Vladimir Ivanov, Wei Li,
  Wenchen Wang, Wenwen Jiang, Wes Bouaziz, Will Constable, Xiaocheng Tang,
  Xiaojian Wu, Xiaolan Wang, Xilun Wu, Xinbo Gao, Yaniv Kleinman, Yanjun Chen,
  Ye~Hu, Ye~Jia, Ye~Qi, Yenda Li, Yilin Zhang, Ying Zhang, Yossi Adi, Youngjin
  Nam, Yu, Wang, Yu~Zhao, Yuchen Hao, Yundi Qian, Yunlu Li, Yuzi He, Zach Rait,
  Zachary DeVito, Zef Rosnbrick, Zhaoduo Wen, Zhenyu Yang, Zhiwei Zhao, and
  Zhiyu Ma. 2024.
\newblock The {Llama} 3 herd of models.
\newblock \emph{arXiv preprint arXiv:2407.21783}.

\bibitem[{He et~al.(2020)He, Wang, Wei, Feng, Mao, and Jiang}]{he2020survey}
Zhiyong He, Zanbo Wang, Wei Wei, Shanshan Feng, Xianling Mao, and Sheng Jiang.
  2020.
\newblock A survey on recent advances in sequence labeling from deep learning
  models.
\newblock \emph{arXiv preprint arXiv:2011.06727}.

\bibitem[{Hewitt et~al.(2024)Hewitt, Liu, Liang, and
  Manning}]{hewitt2024instruction}
John Hewitt, Nelson~F Liu, Percy Liang, and Christopher~D Manning. 2024.
\newblock Instruction following without instruction tuning.
\newblock \emph{arXiv preprint arXiv:2409.14254}.

\bibitem[{Honnibal et~al.(2020)Honnibal, Montani, Van~Landeghem, and
  Boyd}]{spacy}
Matthew Honnibal, Ines Montani, Sofie Van~Landeghem, and Adriane Boyd. 2020.
\newblock \href {https://doi.org/10.5281/zenodo.1212303} {{spaCy}:
  {Industrial}-strength natural language processing in {Python}}.

\bibitem[{Huerta-Enochian and Ko(2024)}]{huerta-enochian-ko-2024-instruction}
Mathew Huerta-Enochian and Seung~Yong Ko. 2024.
\newblock \href {https://doi.org/10.18653/v1/2024.emnlp-main.1267} {Instruction
  fine-tuning: Does prompt loss matter?}
\newblock In \emph{Proceedings of the 2024 Conference on Empirical Methods in
  Natural Language Processing}, pages 22771--22795, Miami, Florida, USA.
  Association for Computational Linguistics.

\bibitem[{Jiang et~al.(2023)Jiang, Sablayrolles, Mensch, Bamford, Chaplot,
  de~las Casas, Bressand, Lengyel, Lample, Saulnier, Lavaud, Lachaux, Stock,
  Scao, Lavril, Wang, Lacroix, and Sayed}]{jiang2023mistral}
Albert~Q. Jiang, Alexandre Sablayrolles, Arthur Mensch, Chris Bamford,
  Devendra~Singh Chaplot, Diego de~las Casas, Florian Bressand, Gianna Lengyel,
  Guillaume Lample, Lucile Saulnier, Lélio~Renard Lavaud, Marie-Anne Lachaux,
  Pierre Stock, Teven~Le Scao, Thibaut Lavril, Thomas Wang, Timothée Lacroix,
  and William~El Sayed. 2023.
\newblock Mistral {7B}.
\newblock \emph{arXiv preprint arXiv:2310.06825}.

\bibitem[{Juki{\'c} and {\v{S}}najder(2024)}]{jukic2024disentangling}
Josip Juki{\'c} and Jan {\v{S}}najder. 2024.
\newblock Disentangling latent shifts of in-context learning through
  self-training.
\newblock \emph{arXiv preprint arXiv:2410.01508}.

\bibitem[{Klamm et~al.(2023)Klamm, Rehbein, and
  Ponzetto}]{klamm-etal-2023-kind}
Christopher Klamm, Ines Rehbein, and Simone~Paolo Ponzetto. 2023.
\newblock \href {https://doi.org/10.18653/v1/2023.findings-eacl.91} {Our kind
  of people? detecting populist references in political debates}.
\newblock In \emph{Findings of the Association for Computational Linguistics:
  EACL 2023}, pages 1227--1243, Dubrovnik, Croatia. Association for
  Computational Linguistics.

\bibitem[{Kwon et~al.(2023)Kwon, Li, Zhuang, Sheng, Zheng, Yu, Gonzalez, Zhang,
  and Stoica}]{kwon2023efficient}
Woosuk Kwon, Zhuohan Li, Siyuan Zhuang, Ying Sheng, Lianmin Zheng, Cody~Hao Yu,
  Joseph~E. Gonzalez, Hao Zhang, and Ion Stoica. 2023.
\newblock Efficient memory management for large language model serving with
  pagedattention.
\newblock In \emph{Proceedings of the ACM SIGOPS 29th Symposium on Operating
  Systems Principles}.

\bibitem[{Lafferty et~al.(2001)Lafferty, McCallum, and
  Pereira}]{lafferty2001conditional}
John~D. Lafferty, Andrew McCallum, and Fernando C.~N. Pereira. 2001.
\newblock Conditional random fields: Probabilistic models for segmenting and
  labeling sequence data.
\newblock In \emph{Proceedings of the Eighteenth International Conference on
  Machine Learning}, ICML '01, page 282–289, San Francisco, CA, USA. Morgan
  Kaufmann Publishers Inc.

\bibitem[{Lhoest et~al.(2021)Lhoest, Villanova~del Moral, Jernite, Thakur, von
  Platen, Patil, Chaumond, Drame, Plu, Tunstall, Davison, {\v{S}}a{\v{s}}ko,
  Chhablani, Malik, Brandeis, Le~Scao, Sanh, Xu, Patry, McMillan-Major, Schmid,
  Gugger, Delangue, Matussi{\`e}re, Debut, Bekman, Cistac, Goehringer, Mustar,
  Lagunas, Rush, and Wolf}]{lhoest-etal-2021-datasets}
Quentin Lhoest, Albert Villanova~del Moral, Yacine Jernite, Abhishek Thakur,
  Patrick von Platen, Suraj Patil, Julien Chaumond, Mariama Drame, Julien Plu,
  Lewis Tunstall, Joe Davison, Mario {\v{S}}a{\v{s}}ko, Gunjan Chhablani,
  Bhavitvya Malik, Simon Brandeis, Teven Le~Scao, Victor Sanh, Canwen Xu,
  Nicolas Patry, Angelina McMillan-Major, Philipp Schmid, Sylvain Gugger,
  Cl{\'e}ment Delangue, Th{\'e}o Matussi{\`e}re, Lysandre Debut, Stas Bekman,
  Pierric Cistac, Thibault Goehringer, Victor Mustar, Fran{\c{c}}ois Lagunas,
  Alexander Rush, and Thomas Wolf. 2021.
\newblock \href {https://doi.org/10.18653/v1/2021.emnlp-demo.21} {Datasets: A
  community library for natural language processing}.
\newblock In \emph{Proceedings of the 2021 Conference on Empirical Methods in
  Natural Language Processing: System Demonstrations}, pages 175--184, Online
  and Punta Cana, Dominican Republic. Association for Computational
  Linguistics.

\bibitem[{Li et~al.(2025{\natexlab{a}})Li, Zhou, Glava{\v{s}}, Korhonen, and
  Vuli{\'c}}]{li2025large}
Chengzu Li, Han Zhou, Goran Glava{\v{s}}, Anna Korhonen, and Ivan Vuli{\'c}.
  2025{\natexlab{a}}.
\newblock Large language models are miscalibrated in-context learners.
\newblock In \emph{Findings of the Association for Computational Linguistics:
  ACL 2025}, pages 11575--11596.

\bibitem[{Li et~al.(2025{\natexlab{b}})Li, Zhang, Do, Yue, and
  Chen}]{li2025longcontext}
Tianle Li, Ge~Zhang, Quy~Duc Do, Xiang Yue, and Wenhu Chen. 2025{\natexlab{b}}.
\newblock \href {https://openreview.net/forum?id=Cw2xlg0e46} {Long-context
  {LLM}s struggle with long in-context learning}.
\newblock \emph{Transactions on Machine Learning Research}.

\bibitem[{Li et~al.(2023)Li, Li, Liu, Xie, Li, Wang, Li, and
  Zhong}]{li2023label}
Zongxi Li, Xianming Li, Yuzhang Liu, Haoran Xie, Jing Li, Fu-lee Wang, Qing Li,
  and Xiaoqin Zhong. 2023.
\newblock Label supervised {LLaMA} finetuning.
\newblock \emph{arXiv preprint arXiv:2310.01208}.

\bibitem[{Liu et~al.(2024{\natexlab{a}})Liu, Lin, Hewitt, Paranjape,
  Bevilacqua, Petroni, and Liang}]{10.1162/tacl_a_00638}
Nelson~F. Liu, Kevin Lin, John Hewitt, Ashwin Paranjape, Michele Bevilacqua,
  Fabio Petroni, and Percy Liang. 2024{\natexlab{a}}.
\newblock \href {https://doi.org/10.1162/tacl_a_00638} {Lost in the middle: How
  language models use long contexts}.
\newblock \emph{Transactions of the Association for Computational Linguistics},
  12:157--173.

\bibitem[{Liu et~al.(2024{\natexlab{b}})Liu, Ye, Xing, and
  Zou}]{10.5555/3692070.3693379}
Sheng Liu, Haotian Ye, Lei Xing, and James Zou. 2024{\natexlab{b}}.
\newblock In-context vectors: {Making} in context learning more effective and
  controllable through latent space steering.
\newblock In \emph{Proceedings of the 41st International Conference on Machine
  Learning}, ICML'24. JMLR.org.

\bibitem[{Liu et~al.(2019)Liu, Ott, Goyal, Du, Joshi, Chen, Levy, Lewis,
  Zettlemoyer, and Stoyanov}]{liu2019roberta}
Yinhan Liu, Myle Ott, Naman Goyal, Jingfei Du, Mandar Joshi, Danqi Chen, Omer
  Levy, Mike Lewis, Luke Zettlemoyer, and Veselin Stoyanov. 2019.
\newblock {RoBERTa}: A robustly optimized {BERT} pretraining approach.
\newblock \emph{arXiv preprint arXiv:1907.11692}.

\bibitem[{Loshchilov and Hutter(2017)}]{loshchilov2016sgdr}
Ilya Loshchilov and Frank Hutter. 2017.
\newblock {SGDR}: Stochastic gradient descent with warm restarts.
\newblock In \emph{International Conference on Learning Representations}.

\bibitem[{Loshchilov and Hutter(2019)}]{loshchilov2017decoupled}
Ilya Loshchilov and Frank Hutter. 2019.
\newblock \href {https://openreview.net/forum?id=Bkg6RiCqY7} {Decoupled weight
  decay regularization}.
\newblock In \emph{International Conference on Learning Representations}.

\bibitem[{Mesquita et~al.(2019)Mesquita, Cannaviccio, Schmidek, Mirza, and
  Barbosa}]{mesquita-etal-2019-knowledgenet}
Filipe Mesquita, Matteo Cannaviccio, Jordan Schmidek, Paramita Mirza, and
  Denilson Barbosa. 2019.
\newblock \href {https://doi.org/10.18653/v1/D19-1069} {{K}nowledge{N}et: A
  benchmark dataset for knowledge base population}.
\newblock In \emph{Proceedings of the 2019 Conference on Empirical Methods in
  Natural Language Processing and the 9th International Joint Conference on
  Natural Language Processing (EMNLP-IJCNLP)}, pages 749--758, Hong Kong,
  China. Association for Computational Linguistics.

\bibitem[{Min et~al.(2022{\natexlab{a}})Min, Lewis, Zettlemoyer, and
  Hajishirzi}]{min-etal-2022-metaicl}
Sewon Min, Mike Lewis, Luke Zettlemoyer, and Hannaneh Hajishirzi.
  2022{\natexlab{a}}.
\newblock \href {https://doi.org/10.18653/v1/2022.naacl-main.201} {{M}eta{ICL}:
  Learning to learn in context}.
\newblock In \emph{Proceedings of the 2022 Conference of the North American
  Chapter of the Association for Computational Linguistics: Human Language
  Technologies}, pages 2791--2809, Seattle, United States. Association for
  Computational Linguistics.

\bibitem[{Min et~al.(2022{\natexlab{b}})Min, Lyu, Holtzman, Artetxe, Lewis,
  Hajishirzi, and Zettlemoyer}]{min-etal-2022-rethinking}
Sewon Min, Xinxi Lyu, Ari Holtzman, Mikel Artetxe, Mike Lewis, Hannaneh
  Hajishirzi, and Luke Zettlemoyer. 2022{\natexlab{b}}.
\newblock \href {https://doi.org/10.18653/v1/2022.emnlp-main.759} {Rethinking
  the role of demonstrations: What makes in-context learning work?}
\newblock In \emph{Proceedings of the 2022 Conference on Empirical Methods in
  Natural Language Processing}, pages 11048--11064, Abu Dhabi, United Arab
  Emirates. Association for Computational Linguistics.

\bibitem[{Mishra et~al.(2022)Mishra, Khashabi, Baral, and
  Hajishirzi}]{mishra-etal-2022-cross}
Swaroop Mishra, Daniel Khashabi, Chitta Baral, and Hannaneh Hajishirzi. 2022.
\newblock \href {https://doi.org/10.18653/v1/2022.acl-long.244} {Cross-task
  generalization via natural language crowdsourcing instructions}.
\newblock In \emph{Proceedings of the 60th Annual Meeting of the Association
  for Computational Linguistics (Volume 1: Long Papers)}, pages 3470--3487,
  Dublin, Ireland. Association for Computational Linguistics.

\bibitem[{Mosbach et~al.(2023{\natexlab{a}})Mosbach, Pimentel, Ravfogel,
  Klakow, and Elazar}]{mosbach2023few}
Marius Mosbach, Tiago Pimentel, Shauli Ravfogel, Dietrich Klakow, and Yanai
  Elazar. 2023{\natexlab{a}}.
\newblock \href {https://doi.org/10.18653/v1/2023.findings-acl.779} {Few-shot
  fine-tuning vs. in-context learning: A fair comparison and evaluation}.
\newblock In \emph{Findings of the Association for Computational Linguistics:
  ACL 2023}, pages 12284--12314, Toronto, Canada. Association for Computational
  Linguistics.

\bibitem[{Mosbach et~al.(2023{\natexlab{b}})Mosbach, Pimentel, Ravfogel,
  Klakow, and Elazar}]{mosbach-etal-2023-shot}
Marius Mosbach, Tiago Pimentel, Shauli Ravfogel, Dietrich Klakow, and Yanai
  Elazar. 2023{\natexlab{b}}.
\newblock \href {https://doi.org/10.18653/v1/2023.findings-acl.779} {Few-shot
  fine-tuning vs. in-context learning: A fair comparison and evaluation}.
\newblock In \emph{Findings of the Association for Computational Linguistics:
  ACL 2023}, pages 12284--12314, Toronto, Canada. Association for Computational
  Linguistics.

\bibitem[{Osborne et~al.(2014)Osborne, Moran, McCreadie, Von~Lunen, Sykora,
  Cano, Ireson, Macdonald, Ounis, He, Jackson, Ciravegna, and
  O{'}Brien}]{osborne-etal-2014-real}
Miles Osborne, Sean Moran, Richard McCreadie, Alexander Von~Lunen, Martin
  Sykora, Elizabeth Cano, Neil Ireson, Craig Macdonald, Iadh Ounis, Yulan He,
  Tom Jackson, Fabio Ciravegna, and Ann O{'}Brien. 2014.
\newblock \href {https://doi.org/10.3115/v1/P14-5007} {Real-time detection,
  tracking, and monitoring of automatically discovered events in social media}.
\newblock In \emph{Proceedings of 52nd Annual Meeting of the Association for
  Computational Linguistics: System Demonstrations}, pages 37--42, Baltimore,
  Maryland. Association for Computational Linguistics.

\bibitem[{Pad{\'o} et~al.(2019)Pad{\'o}, Blessing, Blokker, Dayanik, Haunss,
  and Kuhn}]{pado-etal-2019-sides}
Sebastian Pad{\'o}, Andre Blessing, Nico Blokker, Erenay Dayanik, Sebastian
  Haunss, and Jonas Kuhn. 2019.
\newblock \href {https://doi.org/10.18653/v1/P19-1273} {Who sides with whom?
  towards computational construction of discourse networks for political
  debates}.
\newblock In \emph{Proceedings of the 57th Annual Meeting of the Association
  for Computational Linguistics}, pages 2841--2847, Florence, Italy.
  Association for Computational Linguistics.

\bibitem[{Pradhan et~al.(2013)Pradhan, Moschitti, Xue, Ng, Bj{\"o}rkelund,
  Uryupina, Zhang, and Zhong}]{pradhan-etal-2013-towards}
Sameer Pradhan, Alessandro Moschitti, Nianwen Xue, Hwee~Tou Ng, Anders
  Bj{\"o}rkelund, Olga Uryupina, Yuchen Zhang, and Zhi Zhong. 2013.
\newblock \href {https://aclanthology.org/W13-3516/} {Towards robust linguistic
  analysis using {O}nto{N}otes}.
\newblock In \emph{Proceedings of the Seventeenth Conference on Computational
  Natural Language Learning}, pages 143--152, Sofia, Bulgaria. Association for
  Computational Linguistics.

\bibitem[{Radevski et~al.(2023)Radevski, Gashteovski, Hung, Lawrence, and
  Glava{\v{s}}}]{radevski2023linking}
Gorjan Radevski, Kiril Gashteovski, Chia-Chien Hung, Carolin Lawrence, and
  Goran Glava{\v{s}}. 2023.
\newblock Linking surface facts to large-scale knowledge graphs.
\newblock In \emph{Proceedings of the 2023 Conference on Empirical Methods in
  Natural Language Processing}, pages 7189--7207. Association for Computational
  Linguistics.

\bibitem[{Razumovskaia et~al.(2024{\natexlab{a}})Razumovskaia, Glava{\v{s}},
  Korhonen, and Vuli{\'c}}]{razumovskaia-etal-2024-sqatin}
Evgeniia Razumovskaia, Goran Glava{\v{s}}, Anna Korhonen, and Ivan Vuli{\'c}.
  2024{\natexlab{a}}.
\newblock \href {https://doi.org/10.18653/v1/2024.naacl-long.453} {{SQATIN}:
  Supervised instruction tuning meets question answering for improved dialogue
  {NLU}}.
\newblock In \emph{Proceedings of the 2024 Conference of the North American
  Chapter of the Association for Computational Linguistics: Human Language
  Technologies (Volume 1: Long Papers)}, pages 8195--8211, Mexico City, Mexico.
  Association for Computational Linguistics.

\bibitem[{Razumovskaia et~al.(2022)Razumovskaia, Glava{\v{s}}, Majewska, Ponti,
  and Vuli{\'c}}]{razumovskaia-etal-2022-natural}
Evgeniia Razumovskaia, Goran Glava{\v{s}}, Olga Majewska, Edoardo Ponti, and
  Ivan Vuli{\'c}. 2022.
\newblock \href {https://doi.org/10.18653/v1/2022.acl-tutorials.8} {Natural
  language processing for multilingual task-oriented dialogue}.
\newblock In \emph{Proceedings of the 60th Annual Meeting of the Association
  for Computational Linguistics: Tutorial Abstracts}, pages 44--50, Dublin,
  Ireland. Association for Computational Linguistics.

\bibitem[{Razumovskaia et~al.(2023)Razumovskaia, Vuli{\'c}, and
  Korhonen}]{razumovskaia-etal-2023-transfer}
Evgeniia Razumovskaia, Ivan Vuli{\'c}, and Anna Korhonen. 2023.
\newblock \href {https://doi.org/10.18653/v1/2023.emnlp-main.369}
  {Transfer-free data-efficient multilingual slot labeling}.
\newblock In \emph{Proceedings of the 2023 Conference on Empirical Methods in
  Natural Language Processing}, pages 6041--6055, Singapore. Association for
  Computational Linguistics.

\bibitem[{Razumovskaia et~al.(2024{\natexlab{b}})Razumovskaia, Vuli{\'c}, and
  Korhonen}]{razumovskaia2024analyzing}
Evgeniia Razumovskaia, Ivan Vuli{\'c}, and Anna Korhonen. 2024{\natexlab{b}}.
\newblock Analyzing and adapting large language models for few-shot
  multilingual {NLU}: Are we there yet?
\newblock \emph{arXiv preprint arXiv:2403.01929}.

\bibitem[{Rubin et~al.(2022)Rubin, Herzig, and
  Berant}]{rubin-etal-2022-learning}
Ohad Rubin, Jonathan Herzig, and Jonathan Berant. 2022.
\newblock \href {https://doi.org/10.18653/v1/2022.naacl-main.191} {Learning to
  retrieve prompts for in-context learning}.
\newblock In \emph{Proceedings of the 2022 Conference of the North American
  Chapter of the Association for Computational Linguistics: Human Language
  Technologies}, pages 2655--2671, Seattle, United States. Association for
  Computational Linguistics.

\bibitem[{Team et~al.(2024)Team, Mesnard, Hardin, Dadashi, Bhupatiraju, Pathak,
  Sifre, Rivière, Kale, Love, Tafti, Hussenot, Sessa, Chowdhery, Roberts,
  Barua, Botev, Castro-Ros, Slone, Héliou, Tacchetti, Bulanova, Paterson,
  Tsai, Shahriari, Lan, Choquette-Choo, Crepy, Cer, Ippolito, Reid,
  Buchatskaya, Ni, Noland, Yan, Tucker, Muraru, Rozhdestvenskiy, Michalewski,
  Tenney, Grishchenko, Austin, Keeling, Labanowski, Lespiau, Stanway, Brennan,
  Chen, Ferret, Chiu, Mao-Jones, Lee, Yu, Millican, Sjoesund, Lee, Dixon, Reid,
  Mikuła, Wirth, Sharman, Chinaev, Thain, Bachem, Chang, Wahltinez, Bailey,
  Michel, Yotov, Chaabouni, Comanescu, Jana, Anil, McIlroy, Liu, Mullins,
  Smith, Borgeaud, Girgin, Douglas, Pandya, Shakeri, De, Klimenko, Hennigan,
  Feinberg, Stokowiec, hui Chen, Ahmed, Gong, Warkentin, Peran, Giang, Farabet,
  Vinyals, Dean, Kavukcuoglu, Hassabis, Ghahramani, Eck, Barral, Pereira,
  Collins, Joulin, Fiedel, Senter, Andreev, and Kenealy}]{team2024gemma}
Gemma Team, Thomas Mesnard, Cassidy Hardin, Robert Dadashi, Surya Bhupatiraju,
  Shreya Pathak, Laurent Sifre, Morgane Rivière, Mihir~Sanjay Kale, Juliette
  Love, Pouya Tafti, Léonard Hussenot, Pier~Giuseppe Sessa, Aakanksha
  Chowdhery, Adam Roberts, Aditya Barua, Alex Botev, Alex Castro-Ros, Ambrose
  Slone, Amélie Héliou, Andrea Tacchetti, Anna Bulanova, Antonia Paterson,
  Beth Tsai, Bobak Shahriari, Charline~Le Lan, Christopher~A. Choquette-Choo,
  Clément Crepy, Daniel Cer, Daphne Ippolito, David Reid, Elena Buchatskaya,
  Eric Ni, Eric Noland, Geng Yan, George Tucker, George-Christian Muraru,
  Grigory Rozhdestvenskiy, Henryk Michalewski, Ian Tenney, Ivan Grishchenko,
  Jacob Austin, James Keeling, Jane Labanowski, Jean-Baptiste Lespiau, Jeff
  Stanway, Jenny Brennan, Jeremy Chen, Johan Ferret, Justin Chiu, Justin
  Mao-Jones, Katherine Lee, Kathy Yu, Katie Millican, Lars~Lowe Sjoesund, Lisa
  Lee, Lucas Dixon, Machel Reid, Maciej Mikuła, Mateo Wirth, Michael Sharman,
  Nikolai Chinaev, Nithum Thain, Olivier Bachem, Oscar Chang, Oscar Wahltinez,
  Paige Bailey, Paul Michel, Petko Yotov, Rahma Chaabouni, Ramona Comanescu,
  Reena Jana, Rohan Anil, Ross McIlroy, Ruibo Liu, Ryan Mullins, Samuel~L
  Smith, Sebastian Borgeaud, Sertan Girgin, Sholto Douglas, Shree Pandya,
  Siamak Shakeri, Soham De, Ted Klimenko, Tom Hennigan, Vlad Feinberg, Wojciech
  Stokowiec, Yu~hui Chen, Zafarali Ahmed, Zhitao Gong, Tris Warkentin, Ludovic
  Peran, Minh Giang, Clément Farabet, Oriol Vinyals, Jeff Dean, Koray
  Kavukcuoglu, Demis Hassabis, Zoubin Ghahramani, Douglas Eck, Joelle Barral,
  Fernando Pereira, Eli Collins, Armand Joulin, Noah Fiedel, Evan Senter, Alek
  Andreev, and Kathleen Kenealy. 2024.
\newblock Gemma: Open models based on {Gemini} research and technology.
\newblock \emph{arXiv preprint arXiv:2403.08295}.

\bibitem[{Tjong Kim~Sang and
  Buchholz(2000)}]{tjong-kim-sang-buchholz-2000-introduction}
Erik~F. Tjong Kim~Sang and Sabine Buchholz. 2000.
\newblock \href {https://aclanthology.org/W00-0726/} {Introduction to the
  {C}o{NLL}-2000 shared task chunking}.
\newblock In \emph{Fourth Conference on Computational Natural Language Learning
  and the Second Learning Language in Logic Workshop}.

\bibitem[{Tjong Kim~Sang and
  De~Meulder(2003)}]{tjong-kim-sang-de-meulder-2003-introduction}
Erik~F. Tjong Kim~Sang and Fien De~Meulder. 2003.
\newblock \href {https://aclanthology.org/W03-0419/} {Introduction to the
  {C}o{NLL}-2003 shared task: Language-independent named entity recognition}.
\newblock In \emph{Proceedings of the Seventh Conference on Natural Language
  Learning at {HLT}-{NAACL} 2003}, pages 142--147.

\bibitem[{Touvron et~al.(2023)Touvron, Martin, Stone, Albert, Almahairi,
  Babaei, Bashlykov, Batra, Bhargava, Bhosale, Bikel, Blecher, Ferrer, Chen,
  Cucurull, Esiobu, Fernandes, Fu, Fu, Fuller, Gao, Goswami, Goyal, Hartshorn,
  Hosseini, Hou, Inan, Kardas, Kerkez, Khabsa, Kloumann, Korenev, Koura,
  Lachaux, Lavril, Lee, Liskovich, Lu, Mao, Martinet, Mihaylov, Mishra,
  Molybog, Nie, Poulton, Reizenstein, Rungta, Saladi, Schelten, Silva, Smith,
  Subramanian, Tan, Tang, Taylor, Williams, Kuan, Xu, Yan, Zarov, Zhang, Fan,
  Kambadur, Narang, Rodriguez, Stojnic, Edunov, and Scialom}]{touvron2023llama}
Hugo Touvron, Louis Martin, Kevin Stone, Peter Albert, Amjad Almahairi, Yasmine
  Babaei, Nikolay Bashlykov, Soumya Batra, Prajjwal Bhargava, Shruti Bhosale,
  Dan Bikel, Lukas Blecher, Cristian~Canton Ferrer, Moya Chen, Guillem
  Cucurull, David Esiobu, Jude Fernandes, Jeremy Fu, Wenyin Fu, Brian Fuller,
  Cynthia Gao, Vedanuj Goswami, Naman Goyal, Anthony Hartshorn, Saghar
  Hosseini, Rui Hou, Hakan Inan, Marcin Kardas, Viktor Kerkez, Madian Khabsa,
  Isabel Kloumann, Artem Korenev, Punit~Singh Koura, Marie-Anne Lachaux,
  Thibaut Lavril, Jenya Lee, Diana Liskovich, Yinghai Lu, Yuning Mao, Xavier
  Martinet, Todor Mihaylov, Pushkar Mishra, Igor Molybog, Yixin Nie, Andrew
  Poulton, Jeremy Reizenstein, Rashi Rungta, Kalyan Saladi, Alan Schelten, Ruan
  Silva, Eric~Michael Smith, Ranjan Subramanian, Xiaoqing~Ellen Tan, Binh Tang,
  Ross Taylor, Adina Williams, Jian~Xiang Kuan, Puxin Xu, Zheng Yan, Iliyan
  Zarov, Yuchen Zhang, Angela Fan, Melanie Kambadur, Sharan Narang, Aurelien
  Rodriguez, Robert Stojnic, Sergey Edunov, and Thomas Scialom. 2023.
\newblock Llama 2: Open foundation and fine-tuned chat models.
\newblock \emph{arXiv preprint arXiv:2307.09288}.

\bibitem[{Trautmann(2020)}]{trautmann-2020-aspect}
Dietrich Trautmann. 2020.
\newblock \href {https://aclanthology.org/2020.argmining-1.5/} {Aspect-based
  argument mining}.
\newblock In \emph{Proceedings of the 7th Workshop on Argument Mining}, pages
  41--52, Online. Association for Computational Linguistics.

\bibitem[{Vaswani et~al.(2017)Vaswani, Shazeer, Parmar, Uszkoreit, Jones,
  Gomez, Kaiser, and Polosukhin}]{vaswani2017attention}
Ashish Vaswani, Noam Shazeer, Niki Parmar, Jakob Uszkoreit, Llion Jones,
  Aidan~N Gomez, {\L}ukasz Kaiser, and Illia Polosukhin. 2017.
\newblock Attention is all you need.
\newblock \emph{Advances in neural information processing systems}, 30.

\bibitem[{Wang et~al.(2022)Wang, Li, Yan, Yan, Wang, Wu, and
  Xu}]{wang2022instructionner}
Liwen Wang, Rumei Li, Yang Yan, Yuanmeng Yan, Sirui Wang, Wei Wu, and Weiran
  Xu. 2022.
\newblock Instruction{NER}: A multi-task instruction-based generative framework
  for few-shot {NER}.
\newblock \emph{arXiv preprint arXiv:2203.03903}.

\bibitem[{Wang et~al.(2023)Wang, Zhou, Zu, Xia, Chen, Zhang, Zheng, Ye, Zhang,
  Gui et~al.}]{wang2023instructuie}
Xiao Wang, Weikang Zhou, Can Zu, Han Xia, Tianze Chen, Yuansen Zhang, Rui
  Zheng, Junjie Ye, Qi~Zhang, Tao Gui, et~al. 2023.
\newblock Instruct{UIE}: Multi-task instruction tuning for unified information
  extraction.
\newblock \emph{arXiv preprint arXiv:2304.08085}.

\bibitem[{von Werra et~al.(2020)von Werra, Belkada, Tunstall, Beeching, Thrush,
  Lambert, Huang, Rasul, and Gallouédec}]{vonwerra2022trl}
Leandro von Werra, Younes Belkada, Lewis Tunstall, Edward Beeching, Tristan
  Thrush, Nathan Lambert, Shengyi Huang, Kashif Rasul, and Quentin Gallouédec.
  2020.
\newblock Trl: Transformer reinforcement learning.
\newblock \url{https://github.com/huggingface/trl}.

\bibitem[{Wies et~al.(2023)Wies, Levine, and Shashua}]{wies2023learnability}
Noam Wies, Yoav Levine, and Amnon Shashua. 2023.
\newblock The learnability of in-context learning.
\newblock \emph{Advances in Neural Information Processing Systems},
  36:36637--36651.

\bibitem[{Wiese et~al.(2017)Wiese, Weissenborn, and
  Neves}]{wiese-etal-2017-neural}
Georg Wiese, Dirk Weissenborn, and Mariana Neves. 2017.
\newblock \href {https://doi.org/10.18653/v1/K17-1029} {Neural domain
  adaptation for biomedical question answering}.
\newblock In \emph{Proceedings of the 21st Conference on Computational Natural
  Language Learning ({C}o{NLL} 2017)}, pages 281--289, Vancouver, Canada.
  Association for Computational Linguistics.

\bibitem[{Willard and Louf(2023)}]{willard2023efficient}
Brandon~T Willard and R{\'e}mi Louf. 2023.
\newblock Efficient guided generation for large language models.
\newblock \emph{arXiv preprint arXiv:2307.09702}.

\bibitem[{Ye et~al.(2023)Ye, Wu, Feng, Yu, and Kong}]{ye2023compositional}
Jiacheng Ye, Zhiyong Wu, Jiangtao Feng, Tao Yu, and Lingpeng Kong. 2023.
\newblock Compositional exemplars for in-context learning.
\newblock In \emph{International Conference on Machine Learning}, pages
  39818--39833. PMLR.

\bibitem[{Yin et~al.(2024)Yin, He, Leong, Wang, Yan, Shen, and
  Zhang}]{yin-etal-2024-deeper}
Qingyu Yin, Xuzheng He, Chak~Tou Leong, Fan Wang, Yanzhao Yan, Xiaoyu Shen, and
  Qiang Zhang. 2024.
\newblock \href {https://doi.org/10.18653/v1/2024.findings-emnlp.239} {Deeper
  insights without updates: The power of in-context learning over fine-tuning}.
\newblock In \emph{Findings of the Association for Computational Linguistics:
  EMNLP 2024}, pages 4138--4151, Miami, Florida, USA. Association for
  Computational Linguistics.

\bibitem[{Zhuang et~al.(2025)Zhuang, Singh, Liu, Shang, and
  Gao}]{zhuang2024vector}
Yufan Zhuang, Chandan Singh, Liyuan Liu, Jingbo Shang, and Jianfeng Gao. 2025.
\newblock \href {https://openreview.net/forum?id=xing7dDGh3} {Vector-{ICL}:
  In-context learning with continuous vector representations}.
\newblock In \emph{The Thirteenth International Conference on Learning
  Representations}.

\end{thebibliography}
\bibliographystyle{acl_natbib}

\onecolumn

\appendix

\section{Replication Details}
\label{sec:repl_details}

\begin{table*}
    \centering
    \begin{adjustbox}{width=1.0\linewidth}
    \begin{tabular}{p{2.5cm}  p{4.9cm}  p{4.9cm}}
    \toprule
         \multicolumn{1}{c}{Number of demonstrations} & \multicolumn{1}{c}{Prompt template for training} & \multicolumn{1}{c}{Prompt template for evaluation} \\
         \midrule
         \multicolumn{1}{c}{\multirow{3}{*}{Standard SFT}} & \scriptsize
         <instruction\_tokens\_start>
         
         \#\#\# Instruction: 

\{instruction\}

\#\#\# Options:

\{available\_classes\_for\_task\}

<instruction\_tokens\_end>

<query\_tokens\_start>

\#\#\# Sentence:

\{query\_example\}

\#\#\# Response:

\{query\_response\_completion\}

<query\_tokens\_end> & \scriptsize
         <instruction\_tokens\_start>
         
         \#\#\# Instruction: 

\{instruction\}

\#\#\# Options:

\{available\_classes\_for\_task\}

<instruction\_tokens\_end>

<query\_tokens\_start>

\#\#\# Sentence:

\{query\_example\}

\#\#\# Response:

<query\_tokens\_end>  \\\midrule

\multicolumn{1}{c}{\multirow{3}{*}{\shortstack{SIFT}}} & \scriptsize <instruction\_tokens\_start>
         
         \#\#\# Instruction: 

\{instruction\}

\#\#\# Options:

\{available\_classes\_for\_task\}

<instruction\_tokens\_end>

<context\_tokens\_start>

<demonstration\_\#1\_tokens\_start>

\#\#\# Sentence:

\{example\_\#\_1\}

\#\#\# Response:

\{response\_completion\_\#\_1\}

<demonstration\_\#1\_tokens\_end>

\dots

<demonstration\_\#n\_tokens\_start>

\#\#\# Sentence:

\{example\_\#\_n\}

\#\#\# Response:

\{response\_completion\_\#\_n\}

<demonstration\_\#n\_tokens\_end>

<context\_tokens\_end>

<query\_tokens\_start>

\#\#\# Sentence:

\{query\_example\}

\#\#\# Response:

\{query\_response\_completion\}

<query\_tokens\_end> & \scriptsize 
<instruction\_tokens\_start>
         
         \#\#\# Instruction: 

\{instruction\}

\#\#\# Options:

\{available\_classes\_for\_task\}

<instruction\_tokens\_end>

<context\_tokens\_start>

<demonstration\_\#1\_tokens\_start>

\#\#\# Sentence:

\{example\_\#\_1\}

\#\#\# Response:

\{response\_completion\_\#\_1\}

<demonstration\_\#1\_tokens\_end>

\dots

<demonstration\_\#n\_tokens\_start>

\#\#\# Sentence:

\{example\_\#\_n\}

\#\#\# Response:

\{response\_completion\_\#\_n\}

<demonstration\_\#n\_tokens\_end>

<context\_tokens\_end>

<query\_tokens\_start>

\#\#\# Sentence:

\{query\_example\}

\#\#\# Response:

<query\_tokens\_end> \\
    \bottomrule
    \end{tabular}
    \end{adjustbox}
    \caption{Supervised fine-tuning prompt templates for training and evaluation concerning the number of demonstrations in the prompt}
    \label{tab:token_div}
\end{table*} 
\begin{table*}
    \centering
    \begin{adjustbox}{width=0.8\linewidth}
    \begin{tabular}{p{2cm}  L{5cm}  L{5cm}}
    \toprule
         \multicolumn{1}{c}{Task (dataset)} & \multicolumn{1}{c}{Example for training} & \multicolumn{1}{c}{Example for evaluation} \\
         \midrule
         \multicolumn{1}{l}{\multirow{3}{*}{NER (CoNLL03)}} & \tiny \#\#\# Instruction: 

extract named entities and their type from the input sentence, all entity types are in options
if there are no named entities in the sentence the output should just be 'NA'
if there are multiple extractions from the sentence, the extraction format should be entity\_1\_span:entity\_1\_class;entity\_2\_span:entity\_2\_class;...

\#\#\# Options:

person, location, organization, miscellaneous

\#\#\# Sentence:

LOS ANGELES AT MONTREAL

\#\#\# Response:

LOS ANGELES:organization;MONTREAL:location

\#\#\# Sentence:

EU rejects German call to boycott British lamb .

\#\#\# Response:

EU:organization;German:miscellaneous;British: miscellaneous<eos> & \tiny \#\#\# Instruction: 

extract named entities and their type from the input sentence, all entity types are in options
if there are no named entities in the sentence the output should just be 'NA'
if there are multiple extractions from the sentence, the extraction format should be entity\_1\_span:entity\_1\_class;entity\_2\_span:entity\_2\_class;...

\#\#\# Options:

person, location, organization, miscellaneous

\#\#\# Sentence:

LOS ANGELES AT MONTREAL

\#\#\# Response:

LOS ANGELES:organization;MONTREAL:location

\#\#\# Sentence:

EU rejects German call to boycott British lamb .

\#\#\# Response: \\\midrule
\multicolumn{1}{c}{\multirow{3}{*}{\shortstack{ABAM  \\ (AAC-MW)}}} & \tiny \#\#\# Instruction: 

extract argument aspects and their type from the input sentence, all aspect types are in options
if there are no argument aspects in the sentence the output should just be 'NA'
if there are multiple extractions from the sentence, the extraction format should be aspect\_1\_span:aspect\_1\_class;aspect\_2\_span:aspect\_2\_class;... 

\#\#\# Options:

capital\_vs\_labor, social\_justice/injustice, economic\_impact, prices, low\_skilled, turnover, government, youth\_and\_secondary\_wage\_earners, competition/business\_challenges, motivation/chances, welfare, un/employment\_rate

\#\#\# Sentence:

Reduced Expense for Social Programs : Employees surviving at minimum wage are also often the same people who must rely on additional support of government run social programs to support themselves and their families on such a small amount of income .

\#\#\# Response:

Reduced Expense for Social Programs:welfare;government run social programs:welfare

\#\#\# Sentence:

As the cost of living has jumped by leaps and bounds minimum wage has barely made an impact .

\#\#\# Response:

the cost of living has jumped:social\_justice/injustice<eos> & \tiny \#\#\# Instruction: 

extract argument aspects and their type from the input sentence, all aspect types are in options
if there are no argument aspects in the sentence the output should just be 'NA'
if there are multiple extractions from the sentence, the extraction format should be aspect\_1\_span:aspect\_1\_class;aspect\_2\_span:aspect\_2\_class;... 

\#\#\# Options:

capital\_vs\_labor, social\_justice/injustice, economic\_impact, prices, low\_skilled, turnover, government, youth\_and\_secondary\_wage\_earners, competition/business\_challenges, motivation/chances, welfare, un/employment\_rate

\#\#\# Sentence:

Reduced Expense for Social Programs : Employees surviving at minimum wage are also often the same people who must rely on additional support of government run social programs to support themselves and their families on such a small amount of income .

\#\#\# Response:

Reduced Expense for Social Programs:welfare;government run social programs:welfare

\#\#\# Sentence:

As the cost of living has jumped by leaps and bounds minimum wage has barely made an impact .

\#\#\# Response: \\\midrule
\multicolumn{1}{c}{\multirow{3}{*}{\shortstack{Slot labeling \\ (NLU++)}}} & \tiny \#\#\# Instruction: 

extract slots and their type from the input sentence, all slot label types are in options
if there are no slots in the sentence the output should just be 'NA'
if there are multiple extractions from the sentence, the extraction format should be slot\_1\_span:slot\_1\_class;slot\_2\_span:slot\_2\_class;... 

\#\#\# Options:

time\_from, person\_name, shopping\_category, date\_from, date, number, adults, rooms, amount\_of\_money, kids, people, date\_to, date\_period, time, company\_name, time\_period, time\_to

\#\#\# Sentence:

book a skincare session Saturday at quarter past 5 afternoon

\#\#\# Response:

Saturday:date;quarter past 5 afternoon:time

\#\#\# Sentence:

send 4900 euros to domineque curl after half past 17 today

\#\#\# Response:

4900 euros:amount\_of\_money;domineque curl:person\_name;half past 17:time\_from;today:date<eos> & \tiny \#\#\# Instruction: 

extract slots and their type from the input sentence, all slot label types are in options
if there are no slots in the sentence the output should just be 'NA'
if there are multiple extractions from the sentence, the extraction format should be slot\_1\_span:slot\_1\_class;slot\_2\_span:slot\_2\_class;... 

\#\#\# Options:

time\_from, person\_name, shopping\_category, date\_from, date, number, adults, rooms, amount\_of\_money, kids, people, date\_to, date\_period, time, company\_name, time\_period, time\_to

\#\#\# Sentence:

book a skincare session Saturday at quarter past 5 afternoon

\#\#\# Response:

Saturday:date;quarter past 5 afternoon:time

\#\#\# Sentence:

send 4900 euros to domineque curl after half past 17 today

\#\#\# Response: \\\midrule
\multicolumn{1}{c}{\multirow{3}{*}{\shortstack{SRL \\ (OntoNotes v5.0)}}} & \tiny \#\#\# Instruction: 

extract arguments of the given verb and their semantic roles from the input sentence, all semantic roles are in options
if there are multiple extractions from the sentence, the extraction format should be argument\_1\_span:argument\_1\_role;argument\_2\_span: argument\_2\_role;... 

\#\#\# Options:

ARG0, ARG1, ARG2, ARG3, ARG4, ARGM-ADJ, ARGM-ADV, ARGM-CAU, ARGM-COM, ARGM-DIR, ARGM-DIS, ARGM-EXT, ARGM-GOL, ARGM-LOC, ARGM-MNR, ARGM-MOD, ARGM-NEG, ARGM-PNC, ARGM-PRD, ARGM-PRP, ARGM-TMP, C-ARG0, C-ARG1, C-ARG2, R-ARG0, R-ARG1, V

\#\#\# Sentence:

The wrong things the sinful self does are clear :

\#\#\# Verb:

does

\#\#\# Response:

The wrong things:ARG1;the sinful self:ARG0;does:V

\#\#\# Sentence:

But using foreign - funded banks to scare people has absolutely no meaning with regard to solving the problem .

\#\#\# Verb:

scare

\#\#\# Response:

scare:V;people:ARG1<eos> & \tiny \#\#\# Instruction: 

extract arguments of the given verb and their semantic roles from the input sentence, all semantic roles are in options
if there are multiple extractions from the sentence, the extraction format should be argument\_1\_span:argument\_1\_role;argument\_2\_span: argument\_2\_role;... 

\#\#\# Options:

ARG0, ARG1, ARG2, ARG3, ARG4, ARGM-ADJ, ARGM-ADV, ARGM-CAU, ARGM-COM, ARGM-DIR, ARGM-DIS, ARGM-EXT, ARGM-GOL, ARGM-LOC, ARGM-MNR, ARGM-MOD, ARGM-NEG, ARGM-PNC, ARGM-PRD, ARGM-PRP, ARGM-TMP, C-ARG0, C-ARG1, C-ARG2, R-ARG0, R-ARG1, V

\#\#\# Sentence:

The wrong things the sinful self does are clear :

\#\#\# Verb:

does

\#\#\# Response:

The wrong things:ARG1;the sinful self:ARG0;does:V

\#\#\# Sentence:

But using foreign - funded banks to scare people has absolutely no meaning with regard to solving the problem .

\#\#\# Verb:

scare

\#\#\# Response: \\
    \bottomrule
    \end{tabular}
    \end{adjustbox}
    \caption{Examples from four sequence labeling datasets for $1$-shot ICL and SIFT experiments}
    \label{tab:icl_sift_examples}
\end{table*} 
\subsection{OntoNotes Subsampling}
\label{subsec:srl_processing}

Examples containing semantic roles that appeared less than \num{1000} times in the original 200k-sentence train corpus were removed, reducing the total number of labels from $67$ to $27$. For the training and validation sets, sentences with less than three tokens or more than $50$ tokens were removed. Sampling was then performed from two subsets: sentences with a single predicate and those with multiple predicates. Duplicates based on tokens and BIO tags were dropped. Finally, sentences without predicates were also discarded. Regarding the test set, sentences without predicates and duplicate sentences were dropped. After preprocessing, we randomly sample \num{6000} test set sentences out of \num{26355} with a fixed seed for efficiency reasons.

\subsection{Optimization}
\label{subsec:implementation_details}

For vanilla CLM, we exploit example packing, where we pack short examples in the same input sequence to maximize efficiency during training. We set the maximum input sequence length to \num{1024} for all experiments. To experiment with SRC, we use \textit{DataCollatorForCompletionOnlyLM} implementation from the TRL library \citep{vonwerra2022trl}. With \textit{DataCollatorForCompletionOnlyLM}, we mask all the tokens from the loss function except the $\mathit{QR}$ tokens. For MRC, we implement our own data collator, which masks all tokens from the loss function except the $\mathit{QR}$ and $\mathit{DR}$ tokens (see \Cref{tab:token_div}). These tokens can be easily identified since we provide the collator with a prompt template for training. However, we do not leverage example packing due to the discontinuity of the tokens incurred with token masking through the loss function. If all examples fit into GPU RAM, we use a batch size of eight with four gradient accumulation steps. Otherwise, we reduce the batch size to four and increase the gradient accumulation steps to eight. This approach ensures consistent training performance without exceeding memory limits. We experimented with a higher rank of $r=64$. We got similar results on the validation set, but with substantially more trained parameters per model, which slowed down the overall fine-tuning process. The models are trained in bfloat16 precision. Using this setup, we fit all models into 40GB of GPU memory.

\begin{table}
\centering
\small
\begin{tabular}{lll}
\toprule
Decoder & Base variant identifier & Instruct variant identifier \\ \hline
Gemma & \texttt{google/gemma-7b} & \texttt{google/gemma-1.1-7b-it} \\ \hline
Llama2 & \texttt{meta-llama/Llama-2-7b-hf} & \texttt{meta-llama/Llama-2-7b-chat-hf} \\ \hline
Llama3 & \texttt{meta-llama/Meta-Llama-3-8B} & \texttt{meta-llama/Meta-Llama-3-8B-Instruct} \\ \hline
Llama3.1 & \texttt{meta-llama/Llama-3.1-8B} & \texttt{meta-llama/Llama-3.1-8B-Instruct} \\ \hline
Mistral & \texttt{mistralai/Mistral-7B-v0.1} & \texttt{mistralai/Mistral-7B-Instruct-v0.2} \\
\bottomrule
\end{tabular}
\caption{Decoders and their \textit{Hugging Face Hub} identifiers for \textit{base} and \textit{instruct} model variants}
\label{tab:sift_models}
\end{table}

\subsection{Handling Special Tokens} 
\label{subsec:special_tokens}
We pad the sequences using left-side padding for decoders since the model should be prevented from learning to generate text starting with the padding token, which occurs with right-side padding. Furthermore, we define a padding token for decoders that do not define it explicitly. Padding token is set to either some of the unique reserved tokens or to \texttt{<unk>} token in the case of LLama2-7B and Mistral-7B models. We avoid using the end-of-sequence (\texttt{<eos>}) token as a padding token because the loss is masked out for padding tokens. If the \texttt{<eos>} tokens were used for padding, the model would never learn when to stop generating text. Instead, we explicitly teach the model to recognize when to stop generation by preserving the loss for the \texttt{<eos>} token and including it at the end of each sequence during training (see \Cref{tab:icl_sift_examples}).

\subsection{Causal Mask Removal}
\label{subsec:cm_removal}

In decoder-as-encoder experiments with removed CM, we use the standard softmax token classifier for all tasks except SRL. For the case of the SRL task, we guide the model's prediction based on the \textit{head word} of the verb using a modification of the model architecture. More specifically, the model concatenates the embedding $\mathbf{x}_\mathrm{verb} \in \mathbb{R}^d$ from its embedding matrix with each contextualized token embedding $\mathbf{x}_\mathrm{decoder} \in \mathbb{R}^d$ (the output of the last decoder layer), where $d$ is the decoder's hidden size. The final token representation is a concatenation of the embedding from the last decoder layer and verb embedding: $\mathbf{x} = [\mathbf{x}_\mathrm{decoder}; \mathbf{x}_\mathrm{verb}]$, which is fed to the standard softmax token classifier. For CM removal experiments, we inherit the optimization hyperparameters and the training procedure. We set the maximum sequence length to $256$ in these experiments. 

\subsection{LLM2Vec}
\label{subsec:llm2vec}

In decoder-as-encoder experiments with LLM2Vec, we use the standard softmax token classifier (only the classifier is tuned to be comparable with results from \citet{behnamghaderllm2vec}) for all tasks except SRL. For the case of the SRL task, we guide the model's prediction based on the \textit{head word} of the verb using a modification of the model architecture. More specifically, the model concatenates the embedding $\mathbf{x}_\mathrm{verb} \in \mathbb{R}^d$ from its embedding matrix with each contextualized token embedding $\mathbf{x}_\mathrm{decoder} \in \mathbb{R}^d$ (the output of the last decoder layer), where $d$ is the decoder's hidden size. The final token representation is a concatenation of the embedding from the last decoder layer and verb embedding: $\mathbf{x} = [\mathbf{x}_\mathrm{decoder}; \mathbf{x}_\mathrm{verb}]$, which is fed to the standard softmax token classifier. For LLM2Vec experiments, we inherit the optimization hyperparameters and the training procedure. We set the maximum sequence length to $256$ in these experiments. \textit{Hugging Face Hub} identifiers of the LLM2Vec models we use are: \texttt{McGill-NLP/LLM2Vec-Mistral-7B-Instruct-v2-mntp}, \texttt{McGill-NLP/LLM2Vec-Meta-Llama-3-8B-Instruct-mntp}, \texttt{McGill-NLP/LLM2Vec-Meta-Llama-31-8B-Instruct-mntp}, \texttt{McGill-NLP/LLM2Vec-Llama-2-7b-chat-hf-mntp}.

\subsection{Forming Training Examples}
\label{subsec:forming_training_examples}

The fine-tuning setups are combined with three CLM strategies. Since we always have at least the query example in the training prompt template, we effectively apply CLM to $n+1$ examples from the training set. The demonstrations are sampled from the training set, and we fix the sampling to be dependent on the seed and the query example to ensure that the whole training setup is shared between decoders. \Cref{tab:icl_sift_examples} shows the training examples for the case of $1$-shot SIFT. Analogous to the $1$-shot setup, we prepare the training examples for SIFT experiments with more than one demonstration in the context. Similar holds for standard SFT experiments, where we have zero demonstrations in the context. We also train models with no instruction in the training prompts to experiment with the effect of the instruction on overall performance. 

Inspired by previous work \citep{razumovskaia-etal-2024-sqatin}, we also experimented with the prompts adhering to the question-answering style in our preliminary experiments. In this design, the model is prompted to answer with a span for each span class in the question or provide an \textit{NA} response if there are no spans for the class in the question. However, this design led to worse results on the validation set for ICL and most CLM strategies, introducing long context problems where the prompt lengths grew significantly in size with the increase in the number of classes and length of the context examples. These findings align with previous work, where it has been shown that LLMs struggle to utilize the long contexts \citep{10.1162/tacl_a_00638}. 

\subsection{Forming Evaluation Examples} 
\label{subsec:forming_evaluation_examples}

Examples of prompts for $1$-shot evaluation are shown in \Cref{tab:icl_sift_examples}. Analogous to the $1$-shot setup, we prepare the training examples for evaluation with more than one demonstration in the context. We match the format with the training prompt templates. We omit the \texttt{<eos>} token from the evaluation examples since the model trained to stop with the \texttt{<eos>} token will not continue generating the response. The training and evaluation prompts differ only in the last part, where the evaluation examples do not provide the $\mathit{QR}$ tokens. The model is prompted to complete the output, and the generated output is parsed to obtain IOB2 tags. We sample the demonstrations from the training set as context for evaluations on the examples from the validation and test sets. Importantly, we share the demonstrations in the context between decoders trained under $n$-shot setups and CLM strategies to ensure a fair comparison. Contexts depend on the seed the model was trained on, meaning that we randomly sample new demonstrations for each example, but have four sets of demonstrations in total (due to four seeds). These are shared between ICL, standard SFT, and SIFT experiments for the fairness of evaluation.

\subsection{Evaluation}
\label{subsec:evaluation_details}

We combine the \textit{outlines} and \textit{vLLM} libraries for constrained decoding and accelerated token generation, respectively. \textit{Outlines} library relies on a finite-state machine formulation of the provided regular expression to guide generation for decoder models by operating on the model logits. \textit{vLLM} did not support integration with PEFT methods, so we merge the weights of the QLoRA-trained module with the decoder and perform generation. The merging is performed by summing up the pre-trained weights and scaled LoRA weights. The merged model is saved to the disk, loaded into memory, and transferred to the GPU, where inference is executed. After performing inference, the saved model is removed from the disk, and we keep only the saved LoRA weights, which leaves a negligible memory footprint. Generation setup is shared between all decoder models. We generate tokens using a temperature of $0.1$ and top-p sampling with a threshold of $0.9$. We require the model to generate up to $200$ tokens since there is no query response in any evaluation set longer than $200$ tokens. We heuristically map response spans of decoders to IOB2 tags. We employ greedy span-based matching of predicted spans and their classes with input tokens. We treat all cases in which no predictions are made, all cases where predicted spans do not align with input tokens, or an exception arises during matching due to output generation stochasticity, as if the \textit{O} tag was predicted for every input token. Finally, during parsing, we consider only the first line of the generated response and discard the rest since we notice on the validation set that models trained in few-shot setups tend to overgenerate (overestimate the number of required generated tokens to complete the task) even when we keep the loss on the \texttt{<eos>} token during training. However, they complete the task in the first line of the generated response, so we allow this bias in the parsing of model outputs.

\subsection{Instruction Variations}
\label{subsec:instruction_variations}

Here, we demonstrate the exact instructions that we used for each proposed variation, on the example of NER task:

\begin{exampleblock}{Instruction variation examples for NER task}
\begin{enumerate}
    \item Vanilla \\
\#\#\# Instruction: \\ extract named entities and their type from the input sentence, all entity types are in options
if there are no named entities in the sentence the output should just be ``NA''\\
if there are multiple extractions from the sentence, the extraction format should be entity\_1\_span:entity\_1\_class;entity\_2\_span:entity\_2\_class;... \\
\#\#\# Options:
person, location, organization, miscellaneous
    \item Permuted \\
\#\#\# Instruction: \\
the the no entity their Options: output sentence, be if if entites types and the sentence, all sentence extractions extract be are are organization, format ``NA'' just named in should person, from there entity\_1\_span:entity\_1\_class;entity\_2\_span:entity\_2\_class;...\#\#\# are miscellaneous location, entities should the type multiple from input in options there named the extraction
    \item Nonsense \\
\#\#\# Instruction: 
``The Funniest Joke in the World'' (also ``Joke Warfare'' and ``Killer Joke'') is a Monty Python comedy sketch revolving around a joke that is so funny that anyone who reads or hears it promptly dies from laughter. Ernest Scribbler (Michael Palin), a British ``manufacturer of jokes,'' writes the joke on a piece of paper only to die laughing. His mother (Eric Idle) also immediately dies laughing after reading it, as do the first constables on the scene. Eventually the joke is contained, weaponized, and deployed against Germany during World War II.\footnotemark
\end{enumerate}
\end{exampleblock}
\footnotetext{\footnotesize{Taken from \url{https://en.wikipedia.org/wiki/The_Funniest_Joke_in_the_World}}}

\section{Complementary Results} \label{sec:complementary_res}

\begin{figure}
    \includegraphics[width=\columnwidth]{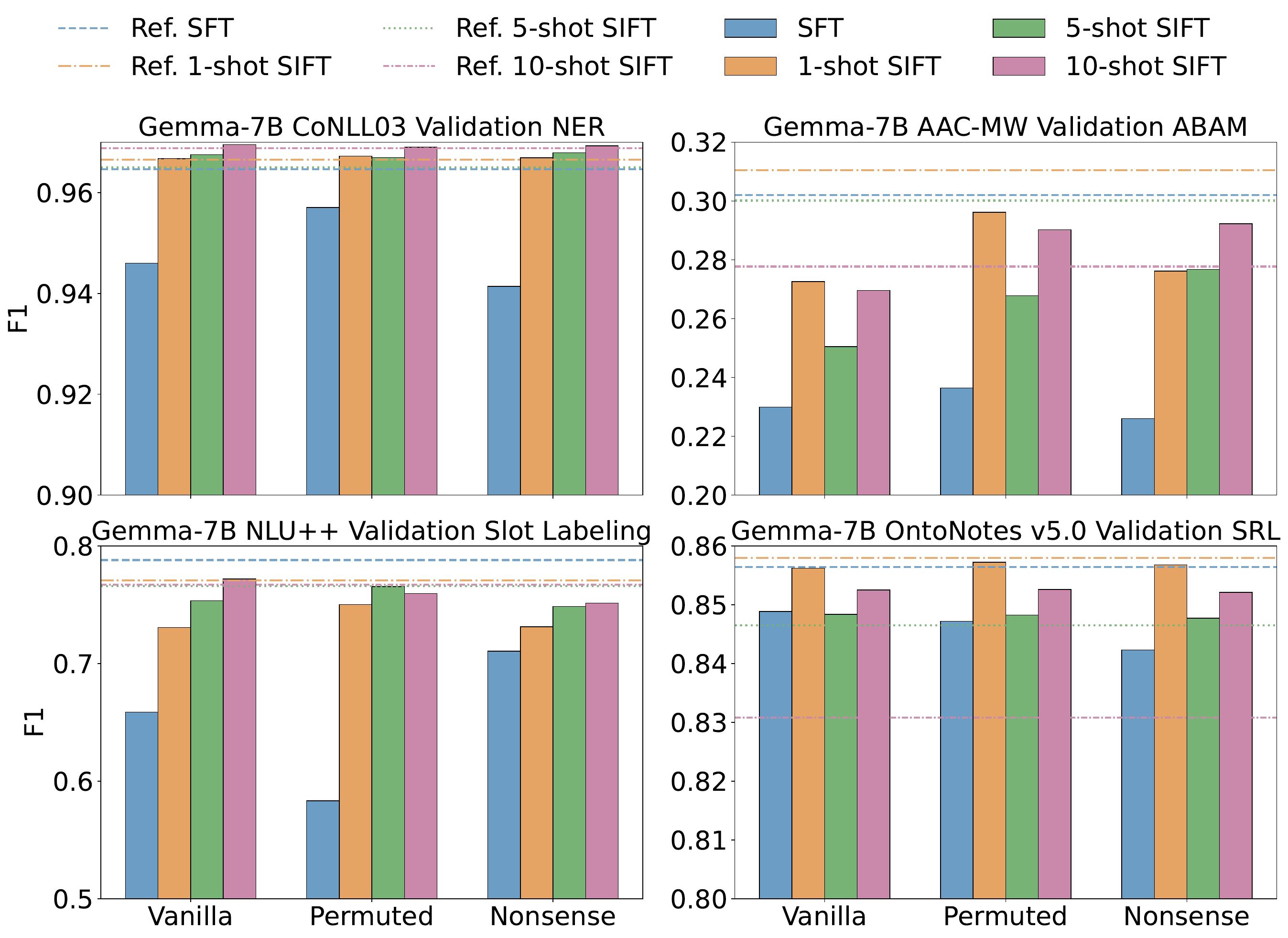}
    \caption{
        Instruction variants applied on the validation set at inference time for Gemma-7B (\textit{base}) trained with MRC and without instructions. The reference (Ref.) denote results for the same models trained with instructions. All results are averages of four runs.
    }
    \label{fig:instruction_effect_gemma}
\end{figure} 
\begin{figure*}
\begin{center}
\includegraphics[width=\textwidth]{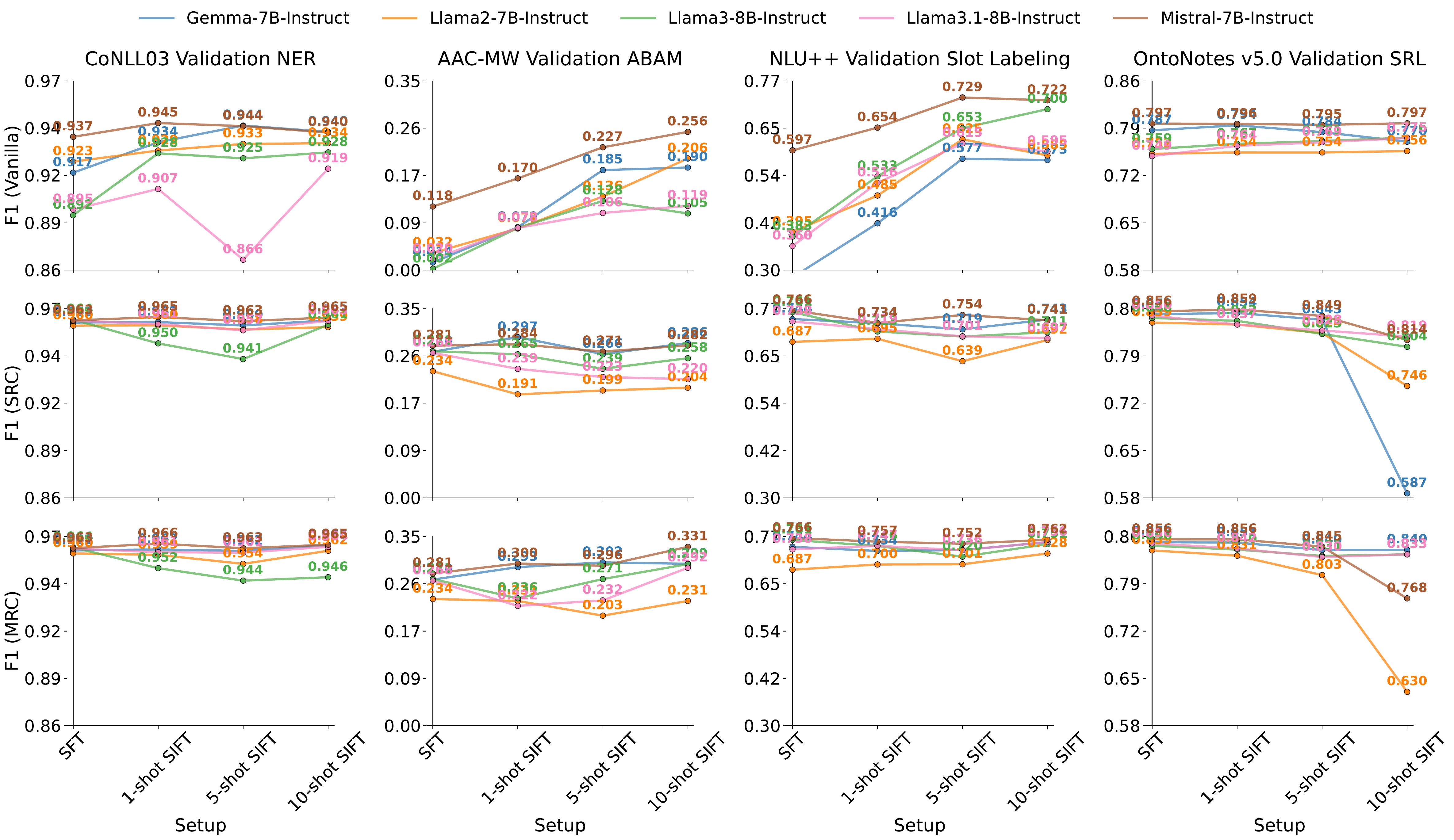}
\caption{Micro F1 scores for five \textit{instruct} variants of decoders on standard SFT and SIFT for a varying number of shots. The models are evaluated with the same number of shots in the context that they used for fine-tuning. The results are given for the validation set on four tasks (left to right) and for three CLM strategies (top to bottom). All results are averages of four runs.}
\label{fig:it_sift_validation_sl_instruct}
\end{center}
\end{figure*}

\end{document}